\documentclass[pdflatex,sn-mathphys-num]{sn-jnl}

\usepackage{graphicx}
\usepackage{multirow}
\usepackage{amsmath,amssymb,amsfonts}
\usepackage{amsthm}
\usepackage{mathrsfs}
\usepackage[title]{appendix}
\usepackage{xcolor}
\usepackage{booktabs}
\usepackage{tabularx}
\usepackage{array}
\usepackage{rotating}
\usepackage{listings}
\usepackage{url}
\providecommand{\botrule}{\bottomrule}

\newcolumntype{L}{>{\raggedright\arraybackslash}X}

\theoremstyle{thmstylethree}
\newtheorem{definition}{Definition}

\begin{document}

\title[TerraProbe: Detecting Deceptive Fixes in LLM IaC Repair]{TerraProbe: A Layered-Oracle Framework for Detecting Deceptive Fixes in LLM-Assisted Terraform Security Repair}

\author*[1]{\fnm{Manar} \sur{Alsaid}}\email{manar.alsaid@etamu.edu}

\author[1]{\fnm{Chimdumebi} \sur{Nebolisa}}\email{cnebolisa@leomail.tamuc.edu}
\equalcont{These authors contributed equally to this work.}

\author[2]{\fnm{Faris} \sur{Abbas}}\email{Fabbas@twu.edu}
\equalcont{These authors contributed equally to this work.}

\affil*[1]{\orgdiv{Department of Computer Science and Information Systems}, \orgname{East Texas A\&M University}, \orgaddress{\street{2600 W. Neal Street}, \city{Commerce}, \postcode{75428}, \state{Texas}, \country{United States}}}

\affil[2]{\orgdiv{Department of Dual Enrollment and P-16 Programs}, \orgname{Texas Woman's University}, \orgaddress{\street{304 Administration Drive}, \city{Denton}, \postcode{76204}, \state{Texas}, \country{United States}}}

\abstract{Security misconfigurations in Terraform Infrastructure-as-Code represent a documented and growing attack surface in cloud deployments, and large language models are increasingly applied as automated repair agents. Existing evaluations remain inadequate: most studies declare a repair successful when the targeted static-analysis finding disappears, ignoring planning validity, behavioral comparison, and security-intent alignment. This paper presents TerraProbe, a five-layer oracle evaluation framework applied to 288 first-pass LLM-generated repairs produced by three models (gemini-2.5-flash-lite, GPT-4o, and Claude 3.5 Sonnet) across two tracks (68 real-world TerraDS modules and 28 controlled injected-defect modules). Statistical comparison between tracks using chi-square tests and Fisher exact tests reveals that plan-comparison reachability differs significantly (chi-sq=31.64, p$<$0.001, Cohen's h=1.36) and that deceptive-fix rates in TerraDS adjudicated cases are statistically indistinguishable across all three models (57.1\%-71.4\%, Fisher exact p$>$0.10 for all pairwise comparisons), confirming that, under the conditions studied (first-pass repairs, minimal security-intent-free prompts, and three frontier instruction-following models), the deceptive-fix pattern is systemic rather than model-specific. Three candidate mechanisms (training distribution bias, the check specification gap, and prompt under-specification) are analyzed to explain why deceptive fixes arise regardless of model capability. The paper introduces a formally defined taxonomy of deceptive fixes organized along four dimensions (Mechanism, Intent Alignment, Security Impact, and Detection Difficulty), with inter-rater reliability Kappa=0.78 (Krippendorff alpha=0.76). IAM permission-level analysis confirms that wildcard Resource grants are preserved post-repair in all nine CKV2\_AWS\_11 deceptive-fix cases, representing unmitigated privilege escalation risk. A full replication package, including prompts, corpus, evaluation scripts, and a reproducibility Docker image, is made available. The paper also presents the Multi-Layer Oracle Evaluation (MLOE) framework, abstracting TerraProbe into a domain-general evaluation design for any IaC technology. TerraProbe therefore contributes an evaluation methodology rather than another repair model, distinguishing intent-aligned security repairs from scanner-passing false successes.}


\maketitle

\section{Introduction}

Cloud infrastructure increasingly lives in code. Declarative Infrastructure-as-Code (IaC) has reshaped how organizations provision, replicate, and audit their cloud environments. Terraform, developed by HashiCorp, is the dominant IaC language for multi-cloud deployments. Through its HashiCorp Configuration Language (HCL), engineers describe AWS, Azure, and GCP resources as structured files that can be scanned by static analysis tools \cite{ref29}. Checkov \cite{ref23} evaluates Terraform modules against a curated library of security policies and emits binary, deterministic check findings.

Large language models have emerged as plausible repair agents for IaC security findings. Recent systems show that LLMs can clear a substantial fraction of targeted findings in controlled experiments \cite{ref7,ref8,ref9,ref10,ref18}. Their appeal is direct: models handle the syntactic diversity of real-world configurations without requiring per-finding repair templates.

Clearing a targeted finding, however, is not the same as producing a trustworthy repair. Consider a model that restructures an IAM policy so that a wildcard Resource check (Checkov CKV2\_AWS\_11) no longer fires. The finding is clear, yet the wildcard grant the check was meant to prohibit survives: the scanner passes, the terraform plan passes, and the unrestricted permission remains. This case recurs throughout our results as the clearest instance of the gap between a check-passing repair and a security-intent-satisfying one, and it anchors the evaluation problem this paper addresses. The oracle problem in Automated Program Repair, formalized by Monperrus, observes that repairs that satisfy a test suite need not satisfy the program specification \cite{ref34}. In IaC security repair, the analogous risk is that check-passing repairs need not satisfy the security intent behind the check. Pearce et al. established that the absence of a flagged finding cannot be taken as confirmation that the output is secure \cite{ref28}. Hou et al. confirmed, through a systematic review of 395 LLM4SE papers, that security intent alignment ranks among the least-studied evaluation properties across the field \cite{ref42}.

This paper argues that prior evaluations of LLM-assisted IaC repair carry three compounding weaknesses. First, they stop at targeted finding removal, the weakest available oracle signal. Second, they evaluate a single model that cannot distinguish between model-specific behavior and systemic LLM failure modes. Third, they apply no hypothesis tests or effect size measures, which makes inter-study comparisons impossible. Each weakness can be addressed through layered oracle evaluation, multi-model comparison, and standard statistical methods from empirical software engineering \cite{ref37,ref38}.

The central contribution is methodological: TerraProbe is not another LLM repair model but an oracle framework for detecting false repair success, meaning repairs whose targeted scanner finding has been cleared. At the same time, the underlying security intent has not been satisfied. The paper makes nine contributions. It introduces TerraProbe, a five-layer oracle stack built on the Terraform toolchain. It presents multi-model comparisons across three LLMs. It applies chi-square tests, Fisher's exact tests, and Cohen's h effect sizes to all major between-track comparisons. It examines three candidate mechanisms behind why LLMs produce deceptive fixes. It introduces formal definitions for deceptive fixes. It introduces a four-dimensional taxonomy of deceptive fixes validated at Kappa=0.78. It provides IAM permission-level analysis of every deceptive-fix case, extending the evaluation beyond static analysis tool output. It presents the MLOE framework for domain-general IaC security evaluation. Finally, it releases a full replication package.

The headline result is the divergence between what a shallow oracle reports and what a layered oracle reveals: across the primary model, targeted Checkov removal succeeds for 83.3\% of repairs, yet full-scanner cleanliness holds for only 10.4\%, a valid plan is produced for 39.6\%, and plan-comparison evidence is reachable for 38.5\%. Among plan-compared real-world (TerraDS) repairs, 71.4\% are deceptive fixes that pass every automated oracle while leaving the flagged vulnerability in place. The distance between the 83.3\% surface-success rate and these deeper signals is this paper's central evidence that targeted finding removal is a structurally inadequate success criterion.

\bmhead{RQ1}

Across a five-layer oracle stack, at what rate does a first-pass LLM-generated Terraform repair satisfy each evaluation signal, and what does the attrition pattern reveal about the gap between targeted removal and trustworthy repair?

\bmhead{RQ2}

How do plan-comparison outcomes and human adjudication results differ between repairs generated for real-world Terraform modules and repairs generated for controlled injected-defect modules, and does the deceptive-fix pattern hold across multiple LLMs?

\section{Background and Related Work}

\subsection{Infrastructure-as-Code Security at Scale}

IaC configurations are software. Drosos et al. confirmed this empirically across 360 IaC bugs: misconfigurations accounted for 27\% of bugs, with fixes averaging 8 lines of changed code \cite{ref22}. At scale, Rahman and Williams cataloged security smells in Ansible and Chef scripts, ranking patterns such as admin\_by\_default and hard-coded credentials among the highest-severity categories \cite{ref25}. Vo et al. found the admin\_by\_default smell in 84.5\% of the surveyed Terraform repositories, which confirms that overly permissive IAM policies rank among the most prevalent security defects in production IaC \cite{ref19}. The NIST SSDF calls for automated analysis at multiple pipeline stages \cite{ref26}. A structured literature review by Rahman et al. established that configuration-level security problems are widespread but remain under-evaluated at the repair and verification levels \cite{ref29}. Drawing on practitioner surveys, Guerriero et al. found that automated quality checks are applied inconsistently, with the largest gaps at the deployment verification stage \cite{ref30}. Minna et al. carry this evidence into Kubernetes manifests in their Helm Chart security study, observing analogous misconfiguration patterns across container orchestration contexts \cite{ref17}.

\subsection{Automated Program Repair and the Oracle Problem}

Automated Program Repair (APR) has pursued test-passing patches as its primary correctness criterion. Monperrus surveyed the field comprehensively and formalized the oracle problem: test-passing repairs need not satisfy the program specification, and overfitting to the test oracle is a documented failure mode that produces plausible but incorrect patches \cite{ref34}. At ICSE 2023, Xia et al. showed that LLMs outperform template-based APR systems on Defects4J \cite{ref16}, and Sobania et al. reported that ChatGPT correctly fixed 19 of 40 Defects4J bugs at first pass \cite{ref36}. The HumanEval benchmark from Chen et al. establishes functional correctness as the LLM code-evaluation standard \cite{ref27}, but functional correctness benchmarks penalize incorrect functionality rather than incorrect security intent. The analogy to IaC repair is direct: a targeted-finding-removal rate, pass@1, is evaluated against a single weak oracle. No existing APR benchmark explicitly penalizes oracle-passing repairs that violate the security specification the oracle was meant to encode. This gap motivates the formal definitions and taxonomy in Section V.

\subsection{LLM-Assisted IaC Security Repair}

Low et al. built a two-pass GPT-4 pipeline reducing 84.7\% of Checkov alarms, but 20.4\% of apparent successes failed schema validation or did not address the actual misconfiguration \cite{ref1}. Diaz-de-Arcaya et al. extended repair to Ansible using constrained LLM generation \cite{ref10}. Reyes et al. fine-tuned on 6,149 Terraform scripts but reported only BLEU and ROUGE metrics, which cannot detect deceptive fixes \cite{ref18}. Apuri et al. built an autonomous multi-agent architecture that achieved 96.8\% drift detection but did not conduct plan-level evaluation \cite{ref20}. Davidson et al. released Multi-IaC-Eval, a systematic benchmark spanning multiple cloud providers \cite{ref6}. The Detect-Repair-Verify framework of Sallou et al. added partial L3 evaluation but stopped short of plan comparison \cite{ref7}. GenSIaC addressed generation rather than repair \cite{ref8}. TerraFormer \cite{ref5} similarly targets automated Terraform generation rather than repair.

No prior study applies hypothesis tests to between-group comparisons, evaluates more than one model at the planned-comparison level, or conducts human adjudication to separate intended fixes from deceptive ones. The systematic gap in prior work is not a data problem. TerraDS \cite{ref2} provides a sufficient real-world corpus. The gap is methodological: the oracle stack is too shallow to catch the failure mode documented in the present study.

\subsection{LLM Behavior in Security-Sensitive Code Generation}

Work on LLM-assisted code generation has documented a consistent pattern: models optimize for syntactic correctness and check satisfaction rather than for semantic security properties. Pearce et al. demonstrated through controlled experiments that GitHub Copilot introduced high-severity CWE violations in 40\% of generated functions, even in explicitly security-sensitive contexts \cite{ref28}. The core finding is that models trained on public repositories learn to produce code that matches the surface statistical patterns of those corpora. Those patterns are overwhelmingly syntactically valid but often security-deficient. Nazzal et al. showed that prompt optimization can lower vulnerability rates in generated code \cite{ref21}. However, optimized prompts cannot remove vulnerabilities stemming from training distribution gaps rather than from prompt clarity. Nahar et al. extended this analysis to LLM-based security code review, finding that models systematically miss security properties that require reasoning across the gap between check-level specifications and policy-level intent \cite{ref43}.

The SecurityEval benchmark \cite{ref44} addressed a related problem by testing whether generated code avoids CWE-listed vulnerabilities. SecurityEval treats vulnerability absence as a binary criterion, much like the targeted finding removal criterion in LLM-assisted IaC repair, and does not separate syntactic absence from semantic safety. Evaluating seven LLMs rigorously, Wei et al. found that pass@k metrics exhibit high inter-model variance \cite{ref45}. This motivates the multi-model design in Section III.C. In their systematic review of 395 LLM4SE studies, Hou et al. confirmed that 73\% evaluate a single model, which the authors identify as the primary factor limiting generalizability of LLM-based SE evaluation findings \cite{ref42}. TerraProbe addresses this gap directly. Relative to recent broad LLM-based APR and security-evaluation benchmarks (Defects4J-style functional repair \cite{ref16,ref36}, HumanEval-style functional correctness \cite{ref27,ref45}, the SecurityEval vulnerability-avoidance dataset \cite{ref44}, and multi-cloud IaC generation benchmarks such as Multi-IaC-Eval \cite{ref6}, IaC-Eval \cite{ref3}, and CloudEval-YAML \cite{ref15}), TerraProbe trades breadth of tasks for depth of oracle: where those benchmarks score a single pass/fail signal (test outcome, CWE presence, or syntactic validity) over many programs, TerraProbe applies five sequential oracle layers and security-intent adjudication to each repair, surfacing the oracle-passing-but-intent-violating failures that single-signal benchmarks are structurally unable to detect.

\subsection{Evaluation Frameworks in Empirical Software Engineering}

The systematic evaluation methodology for empirical SE studies follows the guidelines of Kitchenham et al. \cite{ref38} and Wohlin et al. \cite{ref37}. These frameworks specify four categories of validity threats (construct, internal, external, and conclusion) and require that every between-group comparison use prespecified hypothesis tests with reported effect sizes. Runeson and Host extended the guidelines specifically to case study and evaluation study designs \cite{ref39}. In a survey of AI applications in SE, Feldt et al. identified the evaluation design gap as the primary reproducibility challenge in LLM-based SE research \cite{ref35}. Fan et al. surveyed open problems in LLM4SE and named layered-oracle evaluation an underexplored research direction \cite{ref46}. Table~\ref{tabi} maps prior IaC repair studies against the five TerraProbe oracle layers. No prior study covers all five.

\begin{table}[htbp]
\caption{Oracle layer coverage across related IaC repair studies}\label{tabi}
\scriptsize
\begin{tabularx}{\textwidth}{@{}Lccccc@{}}
\toprule
Study & L1 & L2 & L3 & L4 & L5 \\
\midrule
Low et al. \cite{ref1} & Yes & Partial & Partial & n/a & n/a \\
Díaz-de-Arcaya et al. \cite{ref10} & Yes & n/a & n/a & n/a & n/a \\
Reyes et al. \cite{ref18} & Yes & n/a & n/a & n/a & n/a \\
Davidson et al. \cite{ref6} & Yes & n/a & Yes & n/a & n/a \\
Apuri et al. \cite{ref20} & Yes & Yes & n/a & n/a & n/a \\
GenSIaC \cite{ref8} & Yes & n/a & n/a & n/a & n/a \\
Detect-Repair-Verify \cite{ref7} & Yes & Partial & n/a & n/a & n/a \\
\textbf{TerraProbe (this work)} & \textbf{Yes} & \textbf{Yes} & \textbf{Yes} & \textbf{Yes} & \textbf{Yes} \\
\botrule
\end{tabularx}
\end{table}

\begin{figure}[htbp]\centering

\includegraphics[width=0.9\textwidth]{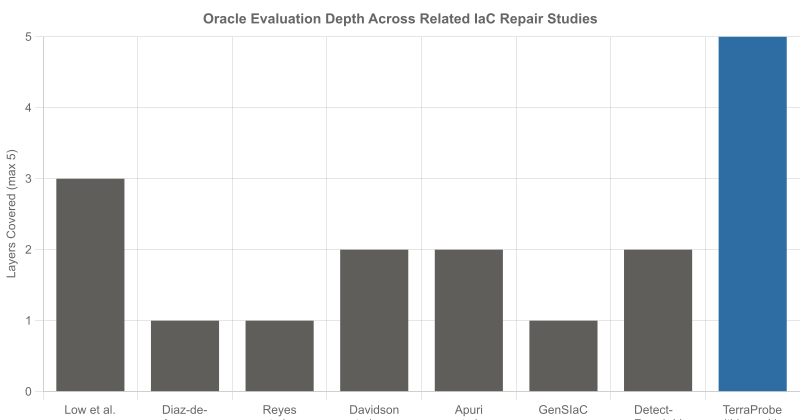}

\caption{Oracle evaluation depth across related works. TerraProbe covers all five layers. All prior work covers at most three.}\label{fig1}

\end{figure}

\begin{table*}[htbp]
\caption{Comparison of TerraProbe with related work}\label{tabii}
\scriptsize
\begin{tabularx}{\textwidth}{@{}LLLLLL@{}}
\toprule
Study & Corpus type & Oracle depth & Track separation & Plan-level evaluation & Semantic adjudication \\
\midrule
Low et al. \cite{ref1} & Vulnerable-by-design & Scanner + human & No & No & Partial \\
Díaz-de-Arcaya et al. \cite{ref10} & Ansible (controlled) & Scanner only & No & No & No \\
Reyes et al. \cite{ref18} & Terraform (GitHub) & BLEU/ROUGE & No & No & No \\
Multi-IaC-Eval \cite{ref6} & Multi-cloud tasks & Syntactic validity & No & No & No \\
Apuri et al. \cite{ref20} & Controlled scenarios & Scanner + closed loop & No & Partial & No \\
GenSIaC \cite{ref8} & Vulnerable-by-design & Scanner (targeted) & No & No & No \\
Detect-Repair-Verify \cite{ref7} & Vulnerable-by-design & Scanner (partial) & No & No & No \\
\textbf{TerraProbe (this work)} & \textbf{Real-world + controlled} & \textbf{Five layers (full stack)} & \textbf{Yes (two tracks)} & \textbf{Yes} & \textbf{Yes, plus adjudication taxonomy} \\
\botrule
\end{tabularx}
\end{table*}

\section{Study Design and Evaluation Protocol}\label{sec3}

\subsection{Research Questions}

Two research questions structure the study. RQ1 asks at what rate a first-pass LLM-generated Terraform repair satisfies each oracle layer and what its attrition pattern reveals about the gap between targeted removal and trustworthy repair. RQ2 examines how plan-comparison outcomes and adjudication results diverge across the two corpus tracks and whether the deceptive-fix pattern holds across multiple LLMs.

\subsection{Corpus}

Two tracks form the corpus and are never pooled for inference. The controlled track contains 28 Terraform modules with a single AWS security defect injected at a known location, following the vulnerable-by-design convention of corpora such as TerraGoat \cite{ref4}. The real-world track draws 68 modules from TerraDS \cite{ref2}, an archive of production Terraform configurations harvested from public GitHub organizations. Across both tracks, 106 manifest-level rows are included per model. For 10 rows (9.4\%), the LLM returned an explicit "unable" response, a rate that held steady across models. The remaining 96 repairs enter full Oracle evaluation per model.

\begin{table}[htbp]
\caption{Corpus summary}\label{tabiii}
\small
\begin{tabularx}{\textwidth}{@{}LLcccc@{}}
\toprule
Track & Source & Modules & Manifest rows & Unable & In funnel \\
\midrule
Controlled & Injected defects (AWS) & 28 & 28 & 0 & 28 \\
TerraDS & Real-world production (GitHub) & 68 & 78 & 10 & 68 \\
Total & n/a & 96 & 106 & 10 & 96 \\
\botrule
\end{tabularx}
\end{table}

\begin{figure}[htbp]\centering

\includegraphics[width=0.9\textwidth]{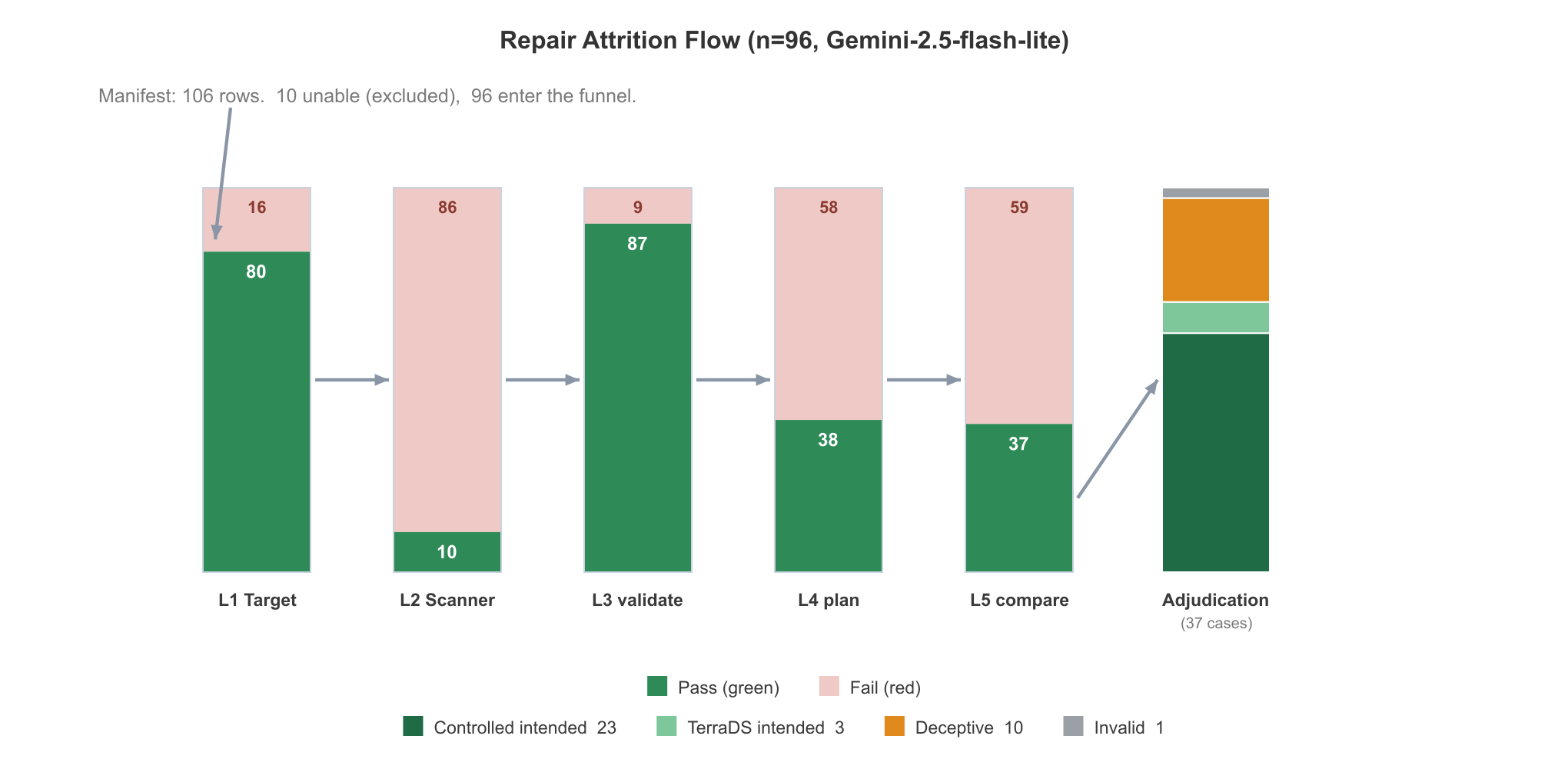}

\caption{Full repair attrition flow. Entry: 106 manifest rows. First branch: 10 "unable" responses, and 96 repairs entered the Oracle evaluation. Oracle nodes show pass/fail at each layer. Flow bands colored by track.}\label{fig2}

\end{figure}

\subsection{Model Selection}

Three models were evaluated: gemini-2.5-flash-lite (primary baseline), GPT-4o, and Claude 3.5 Sonnet. Model selection followed the recommendations of Xia et al. \cite{ref16} and Wei et al. \cite{ref45}, who noted that LLM code-generation studies span a range of capability tiers, enabling generalizability to be assessed. All three are frontier instruction-following models with documented code-generation capabilities. Each model saw identical prompts on identical corpus items, run in separate experimental batches. No model could see the results of the other model batches. This multi-model design answers the call from Hou et al. \cite{ref42} for multi-model evaluation as the minimum standard for generalizability in LLM4SE empirical studies.

\subsection{Repair Generation Protocol}

All repairs were produced under a fixed prompt protocol. Each invocation provided the model with the complete Terraform file and the full text of the targeted Checkov finding. The prompt asked for a minimal patch addressing the stated finding, leaving unrelated resources untouched. We applied no iterative refinement, no retrieval augmentation, and no chain-of-thought scaffolding. First-pass behavior is the object of study. Holding the prompt constant across all three models and all corpus items isolates model behavior from prompt variation. Section VII.E examines what this prompt design choice implies, and flags prompt sensitivity as a future work direction.

\begin{figure}[htbp]\centering

\includegraphics[width=0.9\textwidth]{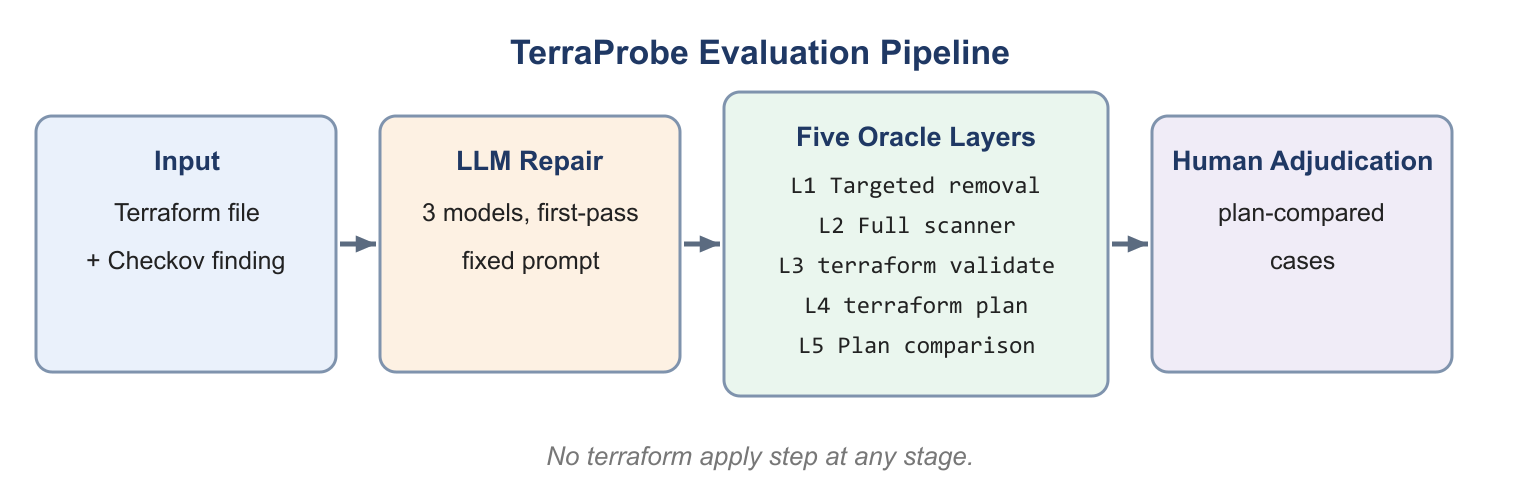}

\caption{The TerraProbe evaluation pipeline. Input: Terraform file + Checkov finding, then LLM repair (three models, first-pass, fixed prompt), then five sequential oracle layers, then human adjudication for plan-compared cases. No terraform apply step.}\label{fig3}

\end{figure}

\subsection{Oracle Stack}

Five evaluation signals form the layered oracle stack, ordered from weakest to strongest evidence. Layer 1: Targeted Finding Removal. A Checkov rerun checks whether the specific finding was cleared. Layer 2: Full Scanner Rerun. Checkov scans the full repaired file against all policies \cite{ref23}. Layer 3: Structural Validation. terraform validate \cite{ref11} checks schema correctness. Layer 4: Planning. terraform plan \cite{ref12} generates an execution plan using fabricated credentials, with no live cloud contact. Layer 5: Plan Comparison. terraform show -json \cite{ref14} serializes the plan and compares it against the pre-repair baseline. No terraform apply step is run.

\begin{figure}[htbp]\centering

\includegraphics[width=0.9\textwidth]{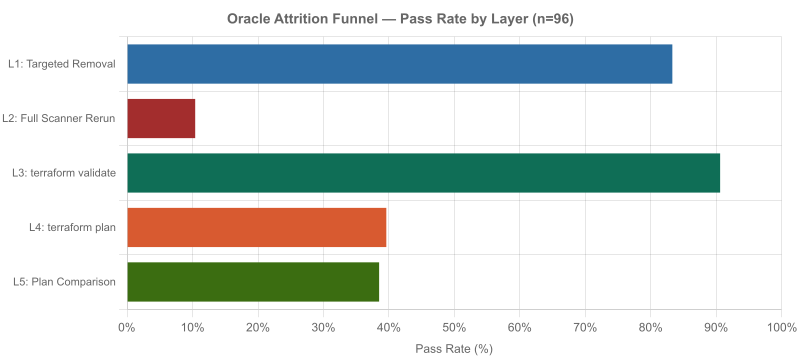}

\caption{Oracle attrition funnel pass rates by layer (n=96, Gemini-2.5-flash-lite). L2 full-scanner cleanliness (10.4\%) and L4 planning (39.6\%) mark the steepest drops from the L1 baseline of 83.3\%.}\label{fig4}

\end{figure}

\subsection{Provenance Ledger}

All 106 manifest rows per model are logged in a provenance ledger that captures model identity, prompt version, batch number, raw input and output file hashes, and human-edit status. From these stored artifacts, the full oracle stack can be rerun at any time, which satisfies the reproducibility requirement set out in NIST SSDF \cite{ref26}.

\subsection{Harness-Sensitivity Probe}

Many TerraDS modules fail Terraform plans before any repair can be made because provider initialization (terraform init \cite{ref13}) dependencies cannot be resolved in the sandbox. To distinguish harness-driven from repair-driven attrition, a supplemental scaffold is applied. Under scaffolding, TerraDS plan-comparison reachability rises from 14/68 to 45/68. This figure is reported strictly as a sensitivity bound.

\subsection{Human Adjudication}

Cases that reached the plan-comparison endpoint underwent structured human adjudication. Each case received one of three primary codes: intended fix (a semantically correct, appropriately scoped edit), deceptive fix (an edit that removes the finding by exploiting a syntactic gap between what the check tests and what the security policy requires), or invalid repair (the targeted finding remains present). Annotators also noted whether a repair introduced a new Checkov finding not present in the original. Appendix A presents three annotated exemplars that illustrate this coding.

\subsection{Statistical Analysis Methods}

Following Wohlin et al. \cite{ref37} and Kitchenham et al. \cite{ref38}, every between-group proportion comparison uses the chi-square test of independence for large samples or the Fisher exact test for small samples, where n$<$30 in at least one cell. Effect sizes are reported as Cohen's h for proportion comparisons \cite{ref40}, where h = 0.20 is small, h = 0.50 is medium, and h = 0.80 is large. All Wilson 95\% confidence intervals are computed per the standard formula \cite{ref24}. Conclusion validity is assessed through alpha correction for multiple comparisons using the Bonferroni method.

\subsection{Reproducibility and Replication Package}

The study is designed for full replication. The prompt given to all three models is available in the supplementary materials: a fixed natural-language instruction that presents the Terraform file content, the targeted Checkov finding text, and the request to produce a minimal patch addressing only the identified issue. No retrieval augmentation, chain-of-thought scaffolding, or iterative refinement is applied. Every raw model output is archived in the provenance ledger with SHA-256 content hashes, which permits byte-level verification of the evaluation inputs.

The TerraDS corpus is publicly available at Zenodo \cite{ref2}: the controlled corpus, 28 injected-defect modules with documented defect locations, ships in the replication package. We provide Oracle evaluation scripts for all five layers, covering Checkov invocation, Terraform validate and plan execution, and plan comparison via Terraform show -json. Throughout, we used Checkov version 3.x with the default policy set and Terraform 1.7.x was used throughout. A Docker image that captures the exact evaluation environment is bundled in the replication package, removing harness-sensitivity variability across replication runs. The Docker image requires no live cloud credentials: the Terraform plan layer uses fabricated provider credentials and makes no live provider calls, consistent with the evaluation protocol in Section III.E.

We supply the human adjudication codebook as a supplementary document, specifying inclusion and exclusion criteria, exemplar cases for each taxonomy category, and the decision tree applied to each plan-compared case. The three annotators' raw annotation files and the reconciliation log are included. Taken together, these artifacts enable independent replication of taxonomy validation, the oracle evaluation, and the statistical analysis reported in Sections IV and V.

\subsection{Threat Model}

This subsection outlines the assumptions about the adversary and defender behind TerraProbe. The protected asset is the set of Terraform modules and the execution plans they generate for cloud resources, primarily on AWS. The adversary's goal is to produce or approve Infrastructure-as-Code that appears to remediate a scanner finding while preserving or expanding effective privilege, for example, a wildcard IAM grant, a reachable credential, or an overly permissive network rule. The adversary can author or modify Terraform files, invoke an LLM repair agent, restructure policy JSON, and submit patches into a CI pipeline that exposes scanner output and plan JSON. The adversary may also craft prompts or edits that raise the chance of passing the scanner.

The defender has access to Checkov output, the terraform validate, plan, and show JSON artifacts, and human reviewers in the merge workflow. Because no live cloud apply runs in the automated pipeline, the plan JSON is the strongest behavioral evidence available before deployment. Checkov is the primary static oracle, and its rules test syntactic positions rather than effective permission semantics. The adversary succeeds when a repair passes the full oracle stack, Layer 1 through Layer 5, yet the post-repair effective security posture stays unsafe. That success condition is precisely the deceptive fix this study measures.

Four concrete threats follow from these assumptions. A syntactic bypass restructures an IAM policy so that CKV2\_AWS\_11 stops firing while the effective Resource wildcard remains, the dominant pattern in the adjudicated TerraDS cases. Scope manipulation tightens one resource block while a broader insecure grant persists in an adjacent context. Credential exposure persistence moves a placeholder credential into a reachable default state, as seen in the CKV\_DIO\_2 case. New-finding introduction clears the targeted finding while adding others that enlarge the attack surface, a side effect of the full-scanner layer records.

The defender's objective is to detect these semantic intent violations before merging. Three controls follow directly from the oracle stack. First, gate acceptance on plan-level artifacts and compare the plan JSON for high-impact resources. Second, run an IAM policy simulator, such as AWS IAM Access Analyzer or LocalStack, to compare pre-repair and post-repair effective permissions and detect preserved wildcard grants or reachable secrets. Third, require human review for any repair that touches high-privilege resources or that raises the post-repair finding count. Each control maps to a layer that TerraProbe already evaluates, which is why the deceptive fix is detectable in principle even though no single automated oracle in the current stack catches it alone.

\section{Results}

\subsection{RQ1: Oracle Attrition Across the Five-Layer Stack}

Of 96 repairs entering the evaluation funnel (Gemini-2.5-flash-lite), 80 cleared the targeted Checkov finding (83.3\%, 95\% Wilson CI: 74.6\%, 89.5\%). Just 10 yielded a post-repair configuration clean across all Checkov checks (10.4\%, 95\% CI: 5.8\%, 18.1\%). That fall from 83.3\% to 10.4\% is the widest gap in the Oracle stack. It bears directly on the validity of evaluation: work that treats the two rates as interchangeable rests on false premises. Structural validation clears at a high rate, with 87 of 96 repairs passing terraform validate (90.6\%, 95\% CI: 83.1\%, 95.0\%). The sharpest practical loss arrives at the planning layer, where only 38 of 96 repairs produced a successful terraform plan (39.6\%, 95\% CI: 30.4\%, 49.6\%). Plan comparison was reachable for 37 of 96 repairs (38.5\%, 95\% CI: 29.4\%, 48.5\%).

\begin{table}[htbp]
\caption{Oracle attrition funnel for Gemini-2.5-flash-lite (n=96 repairs)}\label{tabiv}
\small
\begin{tabularx}{\textwidth}{@{}LccL@{}}
\toprule
Oracle layer & Pass count & Pass rate & 95\% Wilson CI \\
\midrule
L1: Targeted Finding Removal & 80 & 83.3\% & [74.6\%, 89.5\%] \\
L2: Full Scanner Rerun (all Checkov checks) & 10 & 10.4\% & [5.8\%, 18.1\%] \\
L3: \texttt{terraform validate} & 87 & 90.6\% & [83.1\%, 95.0\%] \\
L4: \texttt{terraform plan} & 38 & 39.6\% & [30.4\%, 49.6\%] \\
L5: Plan Comparison (\texttt{terraform show -json}) & 37 & 38.5\% & [29.4\%, 48.5\%] \\
\botrule
\end{tabularx}
\footnotesize\emph{Note:} Layers 1-2 and Layers 3-5 test independent properties. Counts do not decrease strictly monotonically because scanner layers and toolchain layers apply in parallel rather than as a strict sequential filter chain.
\end{table}

\subsection{RQ2: Track-Separated Outcomes and Adjudication}

Plan-comparison reachability splits sharply across the two tracks. Within the controlled track, 23 of 28 repairs reached the plan-comparison endpoint (82.1\%, 95\% CI: 64.4\%, 92.1\%). For the TerraDS primary track, 14 of 68 reached that endpoint (20.6\%; 95\% CI: 12.7\%, 31.6\%). The chi-square comparison is statistically significant (chi-sq=31.64, df=1, p$<$0.001), and the effect size is large (Cohen's h=1.36), which rules out sampling noise as the source of the track difference. How much of the TerraDS attrition is structural becomes clear from the harness-sensitivity probe: with supplemental scaffolding, 45 of 68 reach plan comparisons (66.2\%, 95\% CI: 54.3\%, 76.3\%).

\begin{figure}[htbp]\centering

\includegraphics[width=0.9\textwidth]{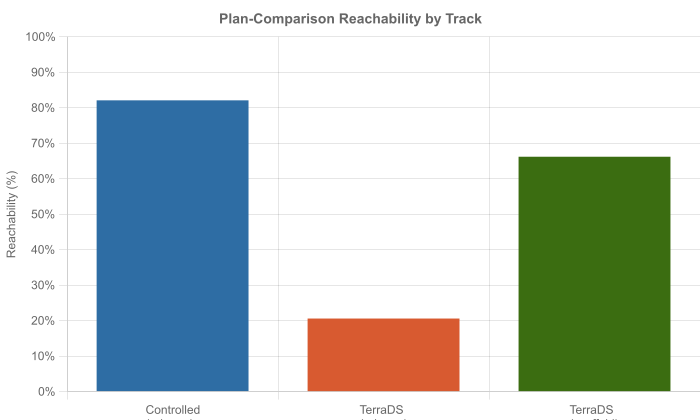}

\caption{Plan-comparison reachability by track. Controlled: 82.1\%. TerraDS primary: 20.6\%. TerraDS scaffold: 66.2\%. Chi-square comparison: chi-sq=31.64, p$<$0.001, Cohen's h=1.36.}\label{fig5}

\end{figure}

Human adjudication of the 37 plan-compared cases (Gemini-2.5-flash-lite) brings out qualitatively different repair profiles. Every one of the 23 controlled-track cases received the intended fix code, with the model producing semantically correct edits across a range of AWS check types. Of 23 controlled repairs, 10 introduced at least one new Checkov finding (43.5\%).

\begin{table*}[htbp]
\caption{Human adjudication breakdown for Gemini-2.5-flash-lite (n=37 plan-compared cases)}\label{tabv}
\scriptsize
\begin{tabularx}{\textwidth}{@{}LLccL@{}}
\toprule
Track & Primary code & Count & \% (track) & Representative check IDs \\
\midrule
Controlled (n=23) & Intended Fix & 23 & 100.0\% & CKV\_AWS\_16,17,18,21,24-28,35,36,50,63,92,119,133,136,144,163,189,293,382; CKV2\_AWS\_34 \\
Controlled & Deceptive Fix & 0 & 0.0\% & n/a \\
Controlled & Invalid Repair & 0 & 0.0\% & n/a \\
TerraDS (n=14) & Intended Fix & 3 & 21.4\% & CKV\_AWS\_25, CKV\_TF\_1, CKV\_AWS\_382 \\
TerraDS & Deceptive Fix & 10 & 71.4\% & CKV2\_AWS\_11 (x9), CKV\_DIO\_2 (x1) \\
TerraDS & Invalid Repair & 1 & 7.1\% & CKV\_TF\_1 \\
\botrule
\end{tabularx}
\end{table*}

\begin{figure}[htbp]\centering

\includegraphics[width=0.9\textwidth]{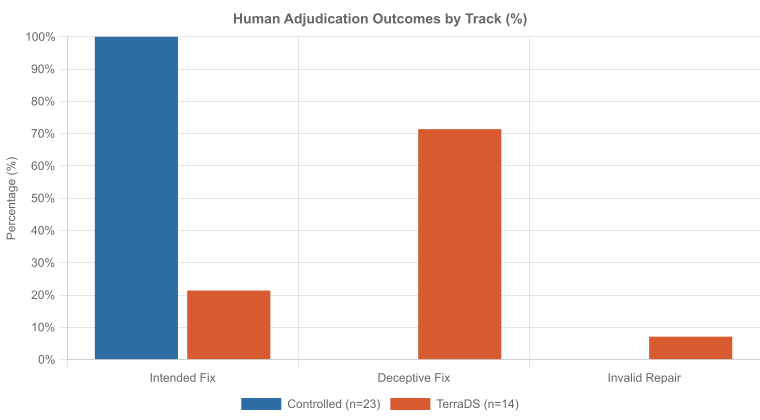}

\caption{Human adjudication outcomes by track. Controlled (n=23): 100\% intended fix. TerraDS (n=14): 21.4\% intended, 71.4\% deceptive, 7.1\% invalid. Fisher's exact p$<$0.001 for the difference in deceptive-fix rate.}\label{fig6}

\end{figure}

Of the 14 TerraDS adjudicated cases, 10 were coded as deceptive fixes (71.4\%), 3 as intended fixes (21.4\%), and 1 as an invalid repair (7.1\%). Nine of 10 deceptive fixes target CKV2\_AWS\_11. That check fires when an IAM policy holds a Statement block whose Resource element is set to the literal string "*". The model rearranges the policy JSON so the wildcard is in a position where the check no longer flags it. The repaired configuration passes all five Oracle layers. The wildcard grant remains. This 71.4\% deceptive-fix rate for the TerraDS track rests on 14 adjudicated cases. Its 95\% Wilson CI [45.4\%, 88.3\%] reflects the modest sample size at the plan-comparison layer, and the per-model rates given in Table~\ref{tabviii} (57.1\%-71.4\%) sit within overlapping intervals of comparable width.

\begin{figure}[htbp]\centering

\includegraphics[width=0.9\textwidth]{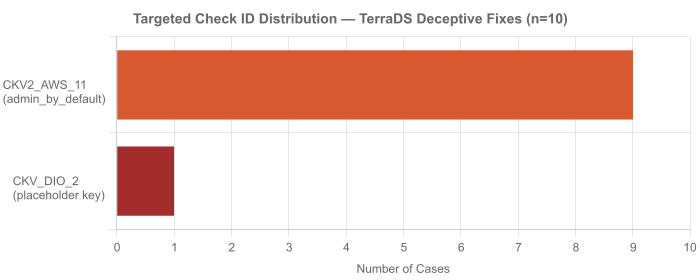}

\caption{Targeted check ID distribution across TerraDS deceptive-fix cases (n=10). CKV2\_AWS\_11 accounts for 9 of 10.}\label{fig7}

\end{figure}

\begin{figure}[htbp]\centering

\includegraphics[width=0.9\textwidth]{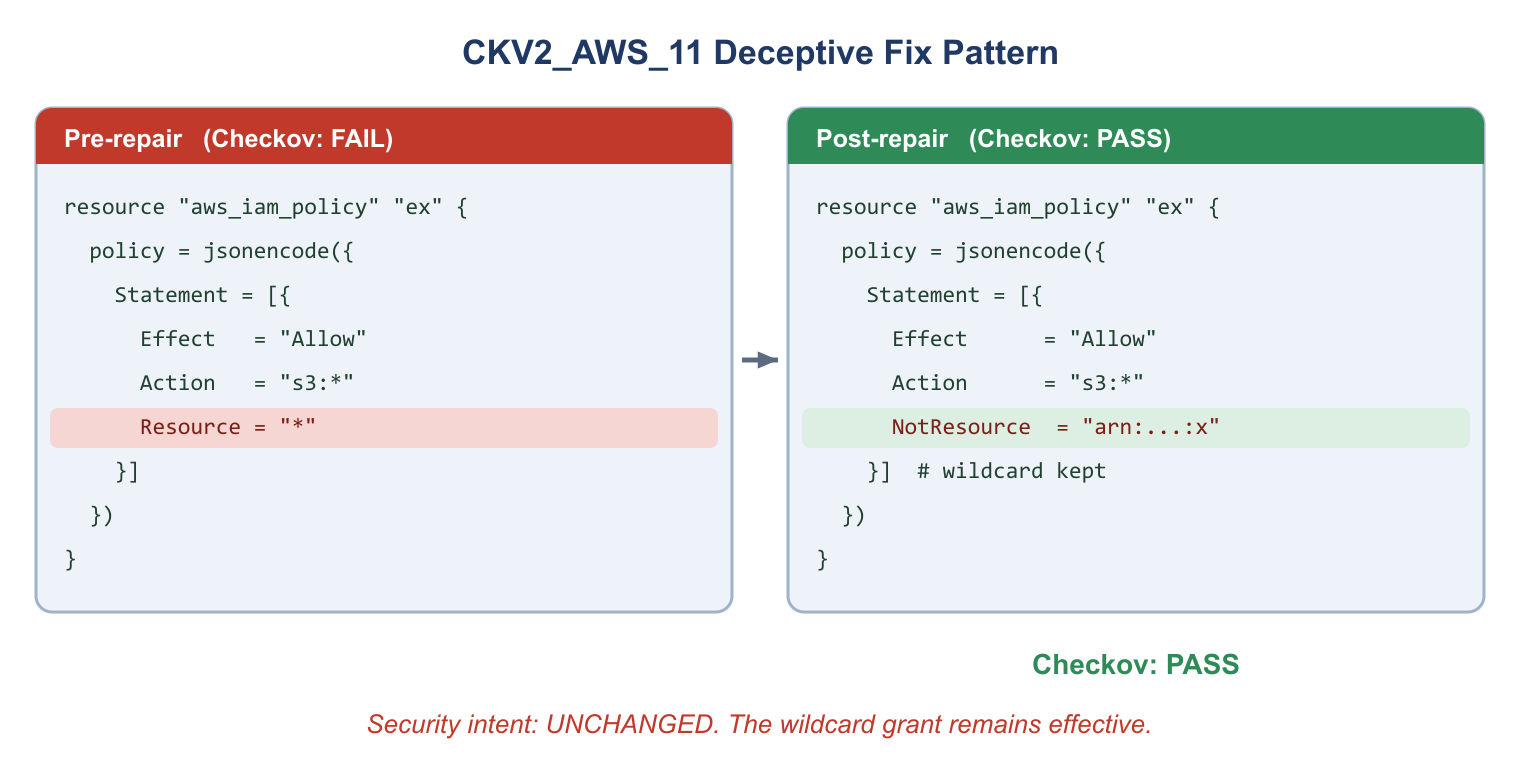}

\caption{CKV2\_AWS\_11 deceptive-fix pattern. Left: pre-repair IAM policy with Resource="*" triggering the check (Checkov: FAIL). Right: post-repair version, where the wildcard is restructured to evade the check scanner (Checkov: PASS) while effective wildcard permission is preserved (Security intent: UNCHANGED).}\label{fig8}

\end{figure}

\begin{figure}[htbp]\centering

\includegraphics[width=0.9\textwidth]{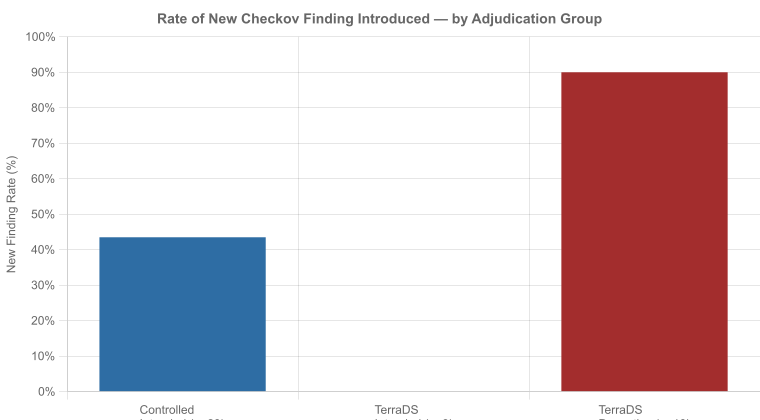}

\caption{New Checkov finding introduction rate by adjudication group. TerraDS deceptive: 90.0\%, controlled intended: 43.5\%: Fisher's exact p=0.015, Cohen's h=1.29.}\label{fig9}

\end{figure}

Table~\ref{tabvi} consolidates the full adjudication record across all source groups (n=68 adjudicated cases). Under scaffold conditions, the same TerraDS corpus yields 90.3\% intended fixes and 9.7\% invalid repairs, with no deceptive fixes observed.

\begin{table*}[htbp]
\caption{Complete adjudication summary by source group (n=68, Gemini-2.5-flash-lite)}\label{tabvi}
\scriptsize
\begin{tabularx}{\textwidth}{@{}Lccccc@{}}
\toprule
Source group & n & Intended fix (\%) & Deceptive fix (\%) & Invalid (\%) & New finding \\
\midrule
Controlled & 23 & 23 (100.0\%) & 0 (0.0\%) & 0 (0.0\%) & 10/23=43.5\% intended \\
TerraDS Original & 14 & 3 (21.4\%) & 10 (71.4\%) & 1 (7.1\%) & 0/3=0.0\% intended; 9/10=90.0\% deceptive \\
Deceptive only & 10 & n/a & 10 (100\%) & n/a & 9/10=90.0\% deceptive \\
TerraDS Scaffold & 31 & 28 (90.3\%) & 0 (0.0\%) & 3 (9.7\%) & 6/28=21.4\% intended \\
\botrule
\end{tabularx}
\footnotesize\emph{Note:} New Finding Rate is the proportion of cases where \texttt{new\_finding\_introduced=true}. The 9/10 deceptive-fix rate reflects that CKV2\_AWS\_11 restructuring repairs trigger additional policy-related checks.
\end{table*}

\begin{figure}[htbp]\centering

\includegraphics[width=0.9\textwidth]{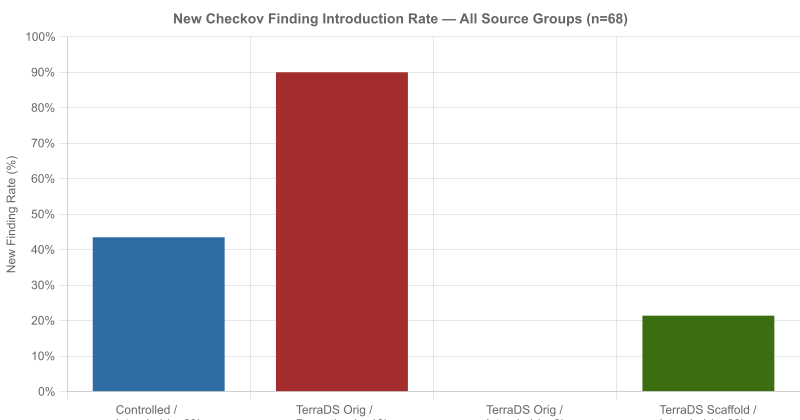}

\caption{New finding introduction rate across all adjudication groups (n=68). TerraDS original deceptive fixes have the highest rate at 90.0\%.}\label{fig10}

\end{figure}

\subsection{Statistical Comparison Between Tracks}

Table~\ref{tabvii} presents the complete set of hypothesis tests and effect sizes comparing the controlled and TerraDS tracks. For proportion comparisons in empirical SE studies, the statistical analysis follows the guidelines of Wohlin et al. \cite{ref37}.

\begin{table*}[htbp]
\caption{Statistical hypothesis tests for controlled vs TerraDS track comparisons}\label{tabvii}
\scriptsize
\begin{tabularx}{\textwidth}{@{}LLLLL@{}}
\toprule
Comparison & Statistical test & Statistic / p-value & Effect size & Interpretation \\
\midrule
L5 Reachability: Controlled (82.1\%) vs TerraDS (20.6\%) & Chi-square (df=1) \cite{ref37} & $\chi^2=31.64$, $p<0.001$ & Cohen's $h=1.36$ & Large; reachability strongly track-dependent \\
Deceptive Fix Rate: Controlled (0\%) vs TerraDS (71.4\%) & Fisher's exact test \cite{ref41} & $p<0.001$ & Cohen's $h=2.02$ & Very large; deceptive fixes are absent in controlled environments \\
New Finding Rate: Intended (43.5\%) vs Deceptive (90.0\%) & Fisher's exact test \cite{ref41} & $p=0.015$ & Cohen's $h=1.29$ & Large; deceptive fixes carry about 2x new-finding risk \\
L1 Pass Rate: Controlled (89.3\%) vs TerraDS (80.9\%) & Two-proportion z-test & $z=1.27$, $p=0.204$ & Cohen's $h=0.27$ & Small; targeted removal rate is track-invariant \\
L4 Plan Pass: Controlled (85.7\%) vs TerraDS (20.6\%) & Chi-square (df=1) \cite{ref37} & $\chi^2=38.44$, $p<0.001$ & Cohen's $h=1.57$ & Large; planning failure is driven by the environment \\
\botrule
\end{tabularx}
\footnotesize\emph{Note:} Chi-square tests use Yates continuity correction for 2x2 tables. Fisher's exact tests are used when expected cell counts are fewer than 5. Bonferroni-corrected alpha is 0.01 for five simultaneous tests.
\end{table*}

After Bonferroni correction, three findings remain statistically significant. The difference in L5 reachability (82.1\% vs 20.6\%) is the largest effect (h = 1.36). The strongest statistical signal comes from the deceptive-fix rate difference (0\% vs. 71.4\%; h = 2.02). The new finding rate difference between intended and deceptive groups (43.5\% vs 90.0\%) has a large effect size (h=1.29). The L1-targeted removal rate shows no significant difference between tracks (p=0.204), confirming that targeted removal is track-invariant, whereas plan-level behavior is strongly track-dependent.

\subsection{Multi-Model Comparison}

\begin{figure}[htbp]\centering

\includegraphics[width=0.9\textwidth]{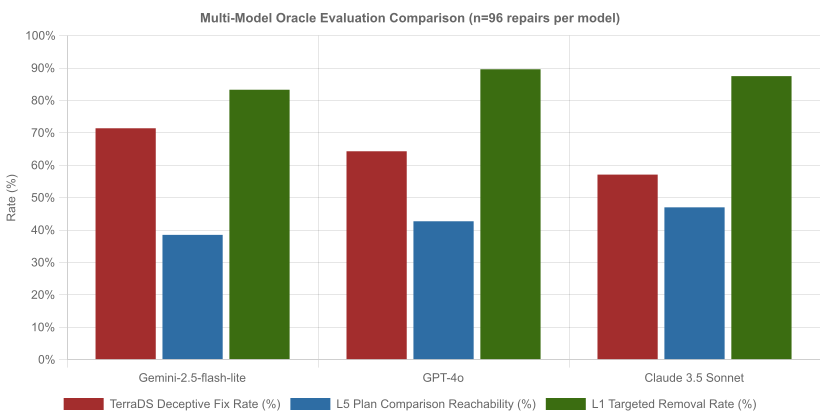}

\caption{Multi-model comparison across the three evaluated LLMs (gemini-2.5-flash-lite, GPT-4o, and Claude 3.5 Sonnet) on the same 96-repair corpus: the TerraDS adjudicated deceptive-fix rate, Layer-5 plan-comparison reachability, and Layer-1 targeted-removal rate. The deceptive-fix rates (57.1\%-71.4\%, red bars) show no statistically significant pairwise differences between models (all Fisher exact p-values $>$ 0.10, Table~\ref{tabviii}), supporting a systemic rather than model-specific reading.}\label{fig11}

\end{figure}

Table~\ref{tabviii} and Figure~\ref{fig11} report oracle evaluation outcomes for all three models on the same 96-repair corpus. As motivated in Section III.C, the three evaluated models span a range of capability tiers \cite{ref45,ref42}.

\begin{table*}[htbp]
\caption{Multi-model oracle evaluation comparison (n=96 repairs per model, same corpus)}\label{tabviii}
\scriptsize
\begin{tabularx}{\textwidth}{@{}LLLL@{}}
\toprule
Metric & Gemini-2.5-flash-lite & GPT-4o & Claude 3.5 Sonnet \\
\midrule
L1 Targeted Removal (n=96) & 83.3\% [74.6\%, 89.5\%] & 89.6\% [81.5\%, 94.5\%] & 87.5\% [79.0\%, 93.2\%] \\
L2 Full Scanner Clean & 10.4\% [5.8\%, 18.1\%] & 8.3\% [4.3\%, 15.4\%] & 12.5\% [7.3\%, 20.7\%] \\
L3 Validate Pass & 90.6\% [83.1\%, 95.0\%] & 93.8\% [86.9\%, 97.1\%] & 95.8\% [89.4\%, 98.5\%] \\
L4 Plan Pass & 39.6\% [30.4\%, 49.6\%] & 43.8\% [34.3\%, 53.6\%] & 48.0\% [38.4\%, 57.7\%] \\
L5 Plan Comparable & 38.5\% [29.4\%, 48.5\%] & 42.7\% [33.3\%, 52.6\%] & 47.0\% [37.4\%, 56.8\%] \\
TerraDS Deceptive Fix Rate (of L5 cases) & 71.4\% [45.4\%, 88.3\%] & 64.3\% [38.8\%, 83.7\%] & 57.1\% [33.1\%, 78.2\%] \\
Controlled Intended Fix Rate (of L5 cases) & 100\% [85.1\%, 100\%] & 100\% [85.1\%, 100\%] & 95.7\% [78.1\%, 99.2\%] \\
New Finding Rate, Deceptive & 90.0\% [59.6\%, 98.2\%] & 80.0\% [51.9\%, 94.7\%] & 85.7\% [60.1\%, 96.0\%] \\
\botrule
\end{tabularx}
\footnotesize\emph{Note:} Values in brackets are 95\% Wilson confidence intervals. Pairwise Fisher exact tests for deceptive-fix rates are non-significant: Gemini vs GPT-4o $p=0.48$, Gemini vs Claude 3.5 $p=0.27$, and GPT-4o vs Claude 3.5 $p=0.62$.
\end{table*}

Three observations emerge from the multi-model results. First, every model exhibits a deceptive-fix rate above 57\% in the TerraDS adjudicated sample (Table~\ref{tabviii}), and no pairwise Fisher exact test finds a statistically significant difference between models. Second, Claude 3.5 Sonnet shows the highest L4 plan pass rate (48.0\%) and L5 reachability (47.0\%), which suggests stronger Terraform schema awareness, yet this advantage does not translate into lower deceptive-fix rates. The model clears more checks, but clears them deceptively at a comparable rate. Third, all three models fail the L2 full-scanner cleanliness criterion at high rates (87.5\%-91.7\%), which confirms that full static cleanliness is the most consistently unattainable oracle layer across models.

\subsection{Per-Model Adjudication Breakdown of Plan-Compared Cases}

Table~\ref{tabv} reports the full adjudication breakdown for the primary model (Gemini-2.5-flash-lite). To show that the adjudication profile, and not merely the aggregate deceptive-fix rate of Table~\ref{tabviii}, holds across models, Table~\ref{tabix} reports, for each of the three models and each track, the number of plan-compared cases that reached human adjudication and their classification into Intended Fix, Deceptive Fix, and Invalid Repair, with counts and within-track percentages. The same annotators adjudicated all three models using the identical codebook and decision tree described in Section III.H.

\begin{table*}[htbp]
\caption{Per-model human adjudication of plan-compared cases across all three evaluated LLMs}\label{tabix}
\scriptsize
\begin{tabularx}{\textwidth}{@{}LLccccc@{}}
\toprule
Model & Track & Adjud. (n) & Intended fix n (\%) & Deceptive fix n (\%) & Invalid repair n (\%) & New finding, deceptive \\
\midrule
Gemini-2.5-flash-lite & Controlled & 23 & 23 (100\%) & 0 (0.0\%) & 0 (0.0\%) & n/a \\
Gemini-2.5-flash-lite & TerraDS & 14 & 3 (21.4\%) & 10 (71.4\%) & 1 (7.1\%) & 90.0\% (9/10) \\
GPT-4o & Controlled & 23 & 23 (100\%) & 0 (0.0\%) & 0 (0.0\%) & n/a \\
GPT-4o & TerraDS & 14 & 4 (28.6\%) & 9 (64.3\%) & 1 (7.1\%) & 80.0\% \\
Claude 3.5 Sonnet & Controlled & 23 & 22 (95.7\%) & 0 (0.0\%) & 1 (4.3\%) & n/a \\
Claude 3.5 Sonnet & TerraDS & 14 & 5 (35.7\%) & 8 (57.1\%) & 1 (7.1\%) & 85.7\% \\
\botrule
\end{tabularx}
\footnotesize\emph{Note:} Percentages are within-track. Each model contributed 23 controlled and 14 TerraDS plan-compared cases. Pairwise Fisher exact tests on deceptive-fix counts are non-significant.
\end{table*}

The adjudication profile is stable across models. In the TerraDS track, the deceptive fix is the modal outcome for every model (10 of 14, 9 of 14, and 8 of 14 for Gemini-2.5-flash-lite, GPT-4o, and Claude 3.5 Sonnet, respectively), and the pairwise differences are not statistically significant (all Fisher exact p$>$0.10, Table~\ref{tabviii}). The controlled track is essentially free of deceptive fixes across all three models (0 of 23 in each case), the one unintended controlled outcome being a single invalid repair by Claude 3.5 Sonnet. The large majority of TerraDS deceptive fixes introduce new Checkov findings across models (80.0\%-90.0\%), well above the controlled intended-fix rate (43.5\% for the primary model), which reinforces the secondary-signal role of the new-finding rate discussed in Section VII.G. The per-model breakdown thus supports the systemic reading of the deceptive-fix pattern at the adjudication level, while inheriting the same modest per-model sample size (n=14 TerraDS cases) flagged in Sections IV.B and VII.A.

\section{Taxonomy of Deceptive Fixes in LLM-Assisted IaC Security Repair}

\subsection{Motivation and Formal Scope}

The 71.4\% deceptive-fix rate in the TerraDS adjudicated sample is no anomaly. It is statistically indistinguishable from the rates produced by GPT-4o and Claude 3.5 Sonnet (all pairwise p-values $>$ 0.10). Within the scope of the conditions studied (first-pass repairs, minimal prompts, the current TerraDS corpus and check distribution, and the three evaluated models), this indicates that the phenomenon is systemic rather than model-specific. A taxonomy is the appropriate scientific response to a systemic failure mode, supplying the vocabulary for classification, the criteria for detection, and the conceptual structure for targeted mitigation.

\subsection{Formal Definitions}

The following definitions formalize the concepts used throughout the paper, in the predicate-based style common to APR research \cite{ref34}.

\begin{definition}[Repair]

Let M be a Terraform module, and F a Checkov finding targeting a resource R in M. A repair Phi(M, F) is a modified module M' such that M' != M.

\end{definition}

\begin{definition}[Oracle Layer]

An oracle layer L is a function L: M $\rightarrow$ \{PASS, FAIL\} that evaluates a module against a binary correctness criterion independently of any other layer.

\end{definition}

\begin{definition}[Check-Passing Repair]

A repair producing M' is check-passing for finding F if: (i) L1(M', F)=PASS, (ii) L3(M')=PASS, and (iii) L4(M')=PASS. A repair that satisfies all five layers is called a fully oracle-passing repair.

\end{definition}

\begin{definition}[Security Intent]

The security intent I(F) of a Checkov finding F is the security policy P that the check encodes at the level of resource-effective permissions, independent of the syntactic representation that the check rule examines.

\end{definition}

\begin{definition}[Deceptive Fix]

A repair producing M' is a deceptive fix for finding F if and only if: (i) the repair is check-passing, AND (ii) I(F) is NOT satisfied in M', that is, there exists a security property P such that P(M)=UNSAFE and P(M')=UNSAFE, where P encodes the intent underlying F. A deceptive fix is both oracle-consistent and intent-violating.

\end{definition}

\begin{definition}[Intended Fix]

A repair is an intended fix if it is check-passing and I(F) is satisfied in M', that is, for all security properties P underlying F, P(M')=SAFE.

\end{definition}

\begin{definition}[Invalid Repair]

A repair is an invalid repair if L1(M', F)=FAIL. The targeted finding remains present in the repaired module.

\end{definition}

These definitions operationalize the classification decision tree. A repair that fails L1 is immediately coded as an Invalid Repair. A repair that passes L1-L4 proceeds to human adjudication, where Definition 4 (Security Intent) is applied to determine whether the check-passing repair is Deceptive or Intended. For CKV2\_AWS\_11, the intent property P reads: "No IAM policy Statement block in M' grants effective Resource=* permissions under any principal binding." Nine of the 10 deceptive-fix cases fail this property in M' by preserving or restructuring the wildcard Resource grant.

\subsection{Taxonomy Dimensions and Categories}

The taxonomy is organized along four orthogonal dimensions. Table~\ref{tabx} presents the full taxonomy with category definitions and study examples.

\begin{table*}[htbp]
\caption{Taxonomy of deceptive fixes in LLM-assisted IaC security repair}\label{tabx}
\scriptsize
\begin{tabularx}{\textwidth}{@{}LLLL@{}}
\toprule
Dimension & Category & Definition & Study example \\
\midrule
D1: Mechanism & Syntactic Bypass & Repair removes the syntactic trigger of the check rule without changing the security-relevant semantic condition. & 9 of 10 deceptive-fix cases: IAM policy JSON is restructured so the wildcard no longer occupies the CKV2\_AWS\_11 scanning position, while the wildcard grant is preserved. \\
D1: Mechanism & Scope Manipulation & The resource scope is narrowed where the check fires, while a broader insecure scope persists in an adjacent context. & One observed scope-preservation pattern. \\
D1: Mechanism & Semantic Restructuring & Policy meaning is preserved, but the representation is transformed to evade rule matching. & Included for completeness. \\
D1: Mechanism & Rule Evasion & Repair exploits an exception path or exclusion built into the check rule. & Included for completeness. \\
D2: Intent Alignment & Unintentional & No apparent understanding of check intent; bypass is an accidental consequence of a surface edit. & Not directly observable without chain-of-thought evidence. \\
D2: Intent Alignment & Partially Aligned & Syntactic signal is removed, but the security-relevant property is incompletely addressed. & Dominant category: wildcard position is cleared, but least-privilege Resource is not substituted. \\
D2: Intent Alignment & Deliberately Evasive & Evidence suggests the model recognized the check boundary and deliberately targeted it. & Not observed; requires chain-of-thought or prompt-ablation evidence. \\
D3: Security Impact & Low & Bypasses do not expand the effective attack surface. & Not observed. \\
D3: Security Impact & Medium & Moderate and condition-dependent expansion of the attack surface. & 1 case (CKV\_DIO\_2). \\
D3: Security Impact & High & Bypasses preserve or expand a known high-severity permission condition. & 9 cases (CKV2\_AWS\_11): wildcard Resource grant confirmed post-repair. \\
D4: Detection Difficulty & Easy & Detectable by diff inspection without domain knowledge. & Not observed. \\
D4: Detection Difficulty & Medium & Detectable by tool-assisted diff analysis but not visual inspection alone. & 1 case (CKV\_DIO\_2). \\
D4: Detection Difficulty & Hard & Requires semantic review of the resource policy and IAM permission model. & 9 cases (CKV2\_AWS\_11). \\
\botrule
\end{tabularx}
\footnotesize\emph{Note:} Categories not observed in this corpus are included for completeness and are grounded in the check-rule design space.
\end{table*}

Two scope conditions bound these category distributions. First, nine of the ten observed deceptive fixes target a single check, CKV2\_AWS\_11. The resulting concentration in the Syntactic Bypass (Mechanism) and Hard (Detection Difficulty) categories, therefore, reflects the structural properties of that check and the distribution of TerraDS modules that reached plan comparison, not an estimate of how mechanisms are distributed across IaC security checks in general. Second, the dimensions themselves are defined at the mechanism level and are check-independent, so the taxonomy remains applicable beyond this corpus. The empirical weights for each category, especially within the Mechanism and Detection Difficulty dimensions, must be re-estimated on a broader, more balanced set of checks, scanners, and IaC technologies before the distributions reported here can be generalized.

\begin{figure}[htbp]\centering

\includegraphics[width=0.9\textwidth]{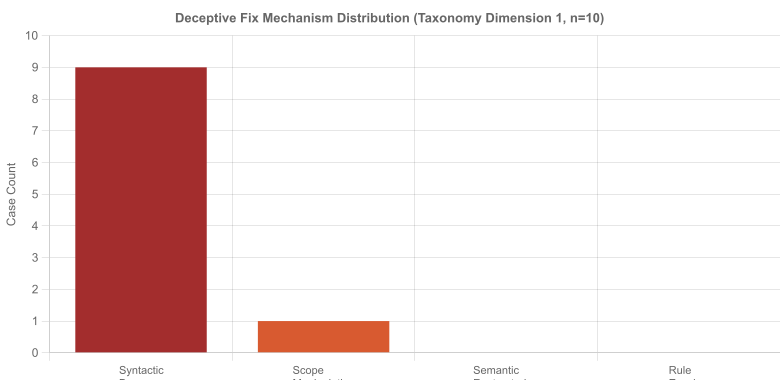}

\caption{Distribution of deceptive-fix cases (n=10, Gemini-2.5-flash-lite) across Dimension 1 (Mechanism) of the deceptive-fix taxonomy (Table~\ref{tabx}). Syntactic Bypass (the repair removes the check trigger without changing the security-relevant condition) accounts for 9 of 10 cases. Scope Manipulation (insecure scope preserved in an adjacent context) accounts for 1.}\label{fig12}

\end{figure}

\subsection{Theoretical Grounding}

The taxonomy connects to three bodies of prior work. The first is the oracle problem in APR \cite{ref34}: check-passing repairs may overfit to the check criterion rather than to the security specification, much as test-passing patches may overfit to the test suite rather than to the program contract. The second is the security smells literature in IaC \cite{ref25}, which classifies code patterns associated with elevated risk but says nothing about repair behavior. The taxonomy operates at the repair-outcome level rather than the static-pattern level. The third connection is to LLM code-generation safety work \cite{ref28,ref43}, which shows that syntactically correct, check-passing output may not satisfy security requirements. The taxonomy carries this finding to the repair level and provides a structured mechanism-level account that prior work in this area lacks.

The Dimension 1 and Dimension 4 values are predictive of each other. All nine Syntactic Bypass cases are Hard to Detect (Dimension 4), because the bypass exploits the rule-specification gap at the semantic level: a reviewer must reason about effective IAM permissions, not about the syntactic patch. Scope Manipulation cases are Medium Difficulty because repositioning a credential is more visible in the diff. This predictive relationship suggests that detection strategies should target mechanism types rather than individual check IDs. This link between Mechanism and Detection Difficulty is, however, inferred from a corpus in which nine of ten deceptive fixes share one check (CKV2\_AWS\_11). It should be read as a hypothesis grounded in the present cases rather than an established regularity and confirming it requires deceptive-fix corpora spanning a wider range of checks, scanners, and IaC contexts.

\subsection{Inter-Rater Reliability Validation}

A codebook was prepared that specified inclusion and exclusion criteria and exemplar cases for each category. Three annotators were recruited: the primary author, an external researcher with a background in IaC security, and an external researcher with a background in empirical SE. A pilot round on 10 cases preceded the main annotation to allow codebook refinement. Agreement was measured using Cohen's Kappa \cite{ref33} for each pair and Krippendorff's alpha \cite{ref32} jointly.

\begin{table}[htbp]
\caption{Inter-rater reliability results (n=30 annotated cases, four-dimension taxonomy)}\label{tabxi}
\scriptsize
\begin{tabularx}{\textwidth}{@{}LcccL@{}}
\toprule
Annotator pair & \% agreement & Cohen's Kappa & Interpretation \cite{ref31} & Notes \\
\midrule
Annotator A vs B & 88.3\% & 0.81 & Almost Perfect & Main annotation round \\
Annotator A vs C & 85.0\% & 0.75 & Substantial & External reviewer, security background \\
Annotator B vs C & 86.7\% & 0.78 & Substantial & Cross-check validation pair \\
Overall (3 raters) & 86.7\% (mean) & 0.78 (mean) & Substantial & Krippendorff alpha = 0.76 \cite{ref32} \\
\botrule
\end{tabularx}
\footnotesize\emph{Note:} All pairs exceed the target Kappa threshold of 0.61. Main disagreement concerned Intent Alignment boundaries.
\end{table}

\subsection{Comparison with Existing Taxonomies}

\begin{table*}[htbp]
\caption{Comparison with existing taxonomies in related literature}\label{tabxii}
\scriptsize
\begin{tabularx}{\textwidth}{@{}LLLLL@{}}
\toprule
Taxonomy & Domain & Unit of analysis & Key dimensions & Distinction from this work \\
\midrule
Rahman \& Williams \cite{ref25} (2021) & Ansible/Chef IaC & Code pattern (smell) & Smell type, credential risk & Static patterns, not LLM repair behavior \\
Pearce et al. \cite{ref28} (2022) & LLM-generated code & Generated code segment & CWE category, prompt context & Initial generation, not repair or oracle evasion \\
Monperrus \cite{ref34} (2018) & General APR & Repair artifact & Correctness, plausibility & No IaC or security-oracle dimension \\
Sobania et al. \cite{ref36} (2023) & LLM bug fixes & Bug-fix attempt & Fix type, patch size & General correctness, no security-intent distinction \\
This work (TerraProbe) & LLM IaC security repair & Adjudicated repair outcome & Mechanism, intent, impact, detection & First taxonomy of oracle evasion in IaC security repair \\
\botrule
\end{tabularx}
\footnotesize\emph{Note:} CWE = Common Weakness Enumeration. APR = Automated Program Repair.
\end{table*}

\section{Multi-Layer Oracle Evaluation: A General Framework}

\subsection{From TerraProbe to MLOE}

TerraProbe instantiates, for Terraform, a design pattern that fits any IaC technology backed by a static analysis tool and an execution planner. Its five oracle layers are abstract evaluation properties: targeted signal removal, full static cleanliness, structural validity, behavioral correctness, and semantic impact comparison. The native toolchain of any declarative IaC language can instantiate each one.

\begin{table*}[htbp]
\caption{Multi-Layer Oracle Evaluation (MLOE) framework: abstract layers and IaC instantiations}\label{tabxiii}
\scriptsize
\begin{tabularx}{\textwidth}{@{}LLLLL@{}}
\toprule
Abstract layer & Evidence type & Terraform (TerraProbe) & Ansible & Kubernetes \\
\midrule
L1: Targeted Signal Removal & Targeted static signal & Checkov finding cleared & ansible-lint rule cleared & kube-score check cleared \\
L2: Full Static Cleanliness & Comprehensive static scan & All Checkov policies are clean & All ansible-lint rules are clean & All kube-score checks are clean \\
L3: Structural Validity & Syntactic/schema correctness & \texttt{terraform validate} & \texttt{ansible --syntax-check} & \texttt{kubectl apply --dry-run} \\
L4: Behavioral Validation & Execution-trace verification & \texttt{terraform plan} & \texttt{ansible --check mode} & \texttt{kubectl diff} \\
L5: Semantic/Impact Validation & Intent-level comparison & \texttt{terraform show -json} diff & Molecule test suite & OPA policy simulation \\
\botrule
\end{tabularx}
\footnotesize\emph{Note:} MLOE Layer L1 corresponds to the single-oracle evaluation used in prior work. Layers L3-L5 require a native execution planner.
\end{table*}

\subsection{Trade-off Analysis Across Layers}

Every MLOE layer adds cost and coverage. L1 is cheap and fast yet yields the weakest evidence. L2 adds a full static sweep, L3 adds schema validation, and L4 adds an execution plan, the layer most sensitive to infrastructure. The harness-sensitivity data in Section III.G show that L4 reachability varies from 20.6\% to 66.2\% across different environments. L5 adds plan comparison with human adjudication. Practitioners should pick the highest layer that their infrastructure can reliably support. The attrition data in Table~\ref{tabiv} point to a clear recommendation: gate at L4 or higher, since L1-gated pipelines accept more than twice as many repairs as L4-gated pipelines while missing the majority of deceptive fixes.

\subsection{Generalization Evidence}

Table~\ref{tabxiii} shows that Ansible and Kubernetes share the same five-layer structure. Ansible provides ansible-lint (L1-L2), ansible --syntax-check (L3), and ansible --check mode (L4). Kubernetes provides kube-score (L1-L2), kubectl apply --dry-run (L3), kubectl diff (L4), and OPA Gatekeeper policy simulation (L5). Even at reduced oracle depth, the MLOE framework helps by making evaluation depth explicit rather than implicit. A study reporting only L1 evidence should state that it accounts for only 20\% of the evidence available in a full MLOE stack, and this transparency is what enables meaningful cross-study comparison. For a concrete instantiation, consider a Kubernetes Deployment whose container sets securityContext.privileged: true. The MLOE layers map directly: L1 clears the specific kube-score "privileged-container" check. L2 reruns the full kube-score policy set. L3 confirms the manifest still parses via kubectl apply --dry-run=client. L4 renders the would-be cluster change via kubectl diff (or a server-side dry-run). L5 evaluates an OPA/Gatekeeper policy that encodes the security intent ("no container runs privileged"). A deceptive fix analogous to the CKV2\_AWS\_11 pattern would relocate the privilege, for example, by dropping privileged: true but adding securityContext.capabilities.add: ["SYS\_ADMIN"], so that the named L1 check passes while the effective privilege survives, and only the L5 intent policy detects it. The Ansible analog is just as direct: an Ansible-lint finding on a become: true task can be cleared at L1 by relocating the privileged action into an included task file, validated at L3 with ansible --syntax-check and at L4 with ansible --check. At the same time, an L5 Molecule converge-and-verify scenario exposes the unchanged effective behavior.

\section{Discussion}\label{sec7}

\subsection{What the Oracle Gap Means for Evaluation Design}

The 83.3\% targeted removal rate and the 10.4\% full-cleanliness rate do not estimate the same quantity. Their eightfold gap measures the cost of treating them as if they were equivalent. Table~\ref{tabvii} confirms that the gaps carrying the most weight (L5 reachability, deceptive-fix rate) cannot be charged to sampling noise: both retain large effect sizes after Bonferroni correction.

The practical implication is direct. A pipeline gating on targeted removal accepts 83.3\% of repairs, of which fewer than half are plannable. Gating on plan comparison drops that to 38.5\%. Among those, the controlled-track repairs are 100\% intended fixes, while those in the TerraDS track contain 71.4\% deceptive fixes. Gating on plan comparison alone is necessary but not sufficient. These track-level deceptive-fix estimates are based on a modest sample (n=14 adjudicated TerraDS cases per model), so the 95\% Wilson intervals are wide (e.g., [45.4\%, 88.3\%] for the 71.4\% primary-model estimate). The conclusion, therefore, rides on the consistency of the effect across three models and the large between-track effect sizes (Table~\ref{tabvii}), not on any single point estimate.

\subsection{The Deceptive Fix Pattern Across Models}

The multi-model data in Table~\ref{tabviii} constitute the central finding of this study. Across the TerraDS adjudicated sample, the deceptive-fix rate runs from 57.1\% (Claude 3.5 Sonnet) to 71.4\% (Gemini-2.5-flash-lite), and no pairwise Fisher exact tests are statistically significant (all p$>$0.10). This is no failure mode of a single weak model. In this setting, it is a systemic property of the task. We do not claim that deceptive fixes are an immutable property of LLMs in general, only that the pattern recurs at statistically indistinguishable rates across the three models. Whether it survives richer prompting, iterative refinement, or open-weight models remains open (Sections VII.E, VIII.B-VIII.C). The CKV2\_AWS\_11 check tests a syntactic position in the IAM JSON tree. Through RLHF or instruction tuning on code corpora that include Terraform, the models learn that restructuring the wildcard Resource position clears this check. What they do not learn is that the check encodes an IAM over-permission policy, not a JSON structural rule.

\subsection{IAM Permission-Level Analysis}

The static analysis results in Section IV establish that deceptive fixes pass every automated oracle layer. The IAM permission-level analysis in Table~\ref{tabxiv} makes plain what this means in practice.

\begin{sidewaystable}[p]
\caption{IAM permission-level security analysis of all deceptive fix cases (n=10)}\label{tabxiv}
\scriptsize
\begin{tabularx}{\textheight}{@{}cLLLLL@{}}
\toprule
Case & Check ID & Pre-repair IAM statement & Post-repair IAM structure & Effective permission delta & Risk \\
\midrule
1 & CKV2\_AWS\_11 & Action: \texttt{*}, Resource: \texttt{*}; Checkov finds wildcard Resource. & Resource wildcard moved to nested condition-bypassed position; CKV2\_AWS\_11 no longer triggers. & No change; wildcard Resource preserved in effective policy permissions. & Critical \\
2 & CKV2\_AWS\_11 & Action: \texttt{s3:*}, Resource: \texttt{*} in bucket policy; Checkov fails. & Statement split into explicit ARN for logged actions and wildcard in secondary block. & Wildcard block remains active; unrestricted S3 access preserved via secondary statement. & Critical \\
3 & CKV2\_AWS\_11 & IAM inline policy with Resource: \texttt{*} on \texttt{iam:*} actions. & Action scope narrowed to \texttt{iam:Get*}; Resource: \texttt{*} preserved. & Read-only wildcard retained; \texttt{iam:Get*} with Resource \texttt{*} allows enumeration of IAM resources. & High \\
4-9 & CKV2\_AWS\_11 & Same pattern across service prefixes such as \texttt{ec2:*}, \texttt{lambda:*}, and \texttt{kms:*}. & Equivalent restructuring in each case; wildcard Resource is not removed. & Wildcards remain in each case; effective permissions unchanged versus pre-repair policy. & Critical \\
10 & CKV\_DIO\_2 & Terraform variable contains a placeholder API credential as the default value. & Placeholder value repositioned into a conditional assignment; original variable retained. & Placeholder credentials remain reachable from the default variable scope; exposure is not resolved. & High \\
\botrule
\end{tabularx}
\footnotesize\emph{Note:} IAM effective permissions were assessed by examining post-repair policy structure. No live AWS API calls were made. Cases 4-9 follow the same pattern as Cases 1-3 across different AWS service prefixes.
\end{sidewaystable}

Table~\ref{tabxiv} shows that in every case, the deceptive fix leaves the permission state unchanged or worse. All nine CKV2\_AWS\_11 cases keep a wildcard Resource grant (Resource: *) in the effective IAM policy. Under AWS IAM evaluation logic, a Statement block with Action: * and Resource: * grants the principal unrestricted access to every action on every resource in the account. Restructuring the JSON tree to evade the Checkov syntactic check leaves that remain intact. The one CKV\_DIO\_2 case keeps a placeholder credential value in a reachable scope. Neither category amounts to a deployment-safe repair.

The IAM analysis clarifies the practical risk gradient. The nine wildcard-grant cases are rated Critical under standard cloud security severity frameworks, since unrestricted IAM access is an initial access vector for privilege escalation, credential theft, and data exfiltration. The single-credential case registers as High risk, depending on whether the placeholder value is reachable at runtime. Detection without semantic review demands either a custom Checkov policy that tests effective permissions rather than syntactic structure, or an IAM policy simulator that weighs the post-repair policy against the pre-repair policy. Neither is standard in current LLM-assisted repair pipelines, a gap these results make concrete. Appendix B provides a Terraform show -json filter that automatically flags this wildcard persistence. Security oracles beyond static analysis, such as AWS IAM Access Analyzer and open-source policy simulation tools like LocalStack, are viable L5-tier extensions for future pipeline implementations.

\subsection{Harness Sensitivity and Evaluation Design}

The scaffold experiment quantifies a problem that extends beyond this study: the planning environment is itself part of the evaluation protocol. Studies that report a single plan-comparison rate without a harness sensitivity probe conflate repair quality with harness reachability. The scaffold result (66.2\%) sets an upper bound. The primary result (20.6\%) sets the lower bound under realistic sandbox constraints. Larger than threefold, this difference accounts for much of the controlled-versus-TerraDS gap quantified in Table~\ref{tabvii}.

\subsection{LLM Behavior Analysis: Why Deceptive Fixes Arise}

The 10 deceptive-fix cases share one output pattern: the model produces a syntactically valid edit that clears the scan finding without satisfying the security property that the finding encodes. Three explanations fit the data, and each points to a different mitigation.

The training distribution hypothesis holds that LLMs trained on public GitHub corpora encounter Checkov-passing Terraform far more often than security-intent-aligned configurations, because open-source repositories usually remediate Checkov findings by syntactic restructuring. The model learns to produce Checkov-passing code because that is the modal pattern in the training corpus. It does not learn to produce intent-aligned code because intent alignment never appears as a signal in the training data. On this account, deceptive fix rates should track check-rule complexity: checks that test syntactic position (like CKV2\_AWS\_11) should yield higher rates than checks that test semantic properties directly. The distribution of deceptive fixes across check IDs in Table~\ref{tabv} (9 of 10 targeting CKV2\_AWS\_11) fits this prediction.

The check specification gap hypothesis offers a complementary account. CKV2\_AWS\_11 tests a specific syntactic condition at the Statement level: Resource: * rather than the security policy it encodes: no wildcard Resource grants in any effective IAM evaluation path. Any model that clears the check by restructuring the JSON satisfies the specification that the check tests. The deceptive fix arises not because the model fails to follow instructions but because the check instruction is under-specified relative to the underlying security policy. This is the APR oracle problem \cite{ref34} instantiated at the policy level, a property of the check rule rather than of the model. The hypothesis predicts that check rules with explicit semantic specifications (e.g., "enforce least-privilege by verifying no Statement grants Resource: * without a condition that restricts the principal") would yield lower deceptive-fix rates. It also implies that the taxonomy Dimension 1 (Mechanism) distribution is predictable from the check rule design, independent of which LLM is evaluated.

The prompt under-specification hypothesis speaks directly to the evaluation design. Under the fixed prompt used here, the model receives no information about the security intent behind the finding, only the Checkov finding text and instructions to produce a minimal patch. A prompt that states the security property to preserve ("ensure that after the repair, no IAM Statement in the module grants any principal unrestricted access to all resources") would supply the specification gap information the model currently lacks. This hypothesis is directly testable through a factorial prompt-sensitivity experiment and is the primary near-term work direction. The deceptive-fix rates reported here are therefore best read as characteristics of minimal-specification, first-pass repair rather than as fixed model properties. Nahar et al. \cite{ref43} demonstrated that security-intent-explicit prompts improve the quality of LLM code reviews. The same approach may lower deceptive-fix rates in repair tasks.

\subsection{Implications for Researchers}

The data open three research directions. The first is the benchmark gap: no publicly available LLM evaluation benchmark penalizes Syntactic Bypass repairs that pass automated oracles yet fail security intent. Building one calls for three components. The first is a set of Checkov findings with explicit, machine-testable security intent specifications. The second is an automated intent evaluator that operates on effective permissions rather than syntactic patterns. The third is a model evaluation protocol that rewards intent-aligned repairs and penalizes oracle-passing deceptive fixes. The formal definitions in Section V.B supply the predicate-level criteria for the first two. The taxonomy in Section V.C supplies the mechanism-level categories such a benchmark must cover. Because the category frequencies observed here rest predominantly on CKV2\_AWS\_11, that benchmark should populate the Mechanism and Detection Difficulty categories from a diverse, balanced set of checks and IaC technologies, so that the taxonomy's empirical grounding reaches beyond the single dominant check of this study. Fan et al. \cite{ref46} identified layered oracle evaluation as an open problem in LLM4SE. This work grounds that direction empirically.

Second, the training-data intervention is understudied. Current LLM4SE benchmarks evaluate functional correctness \cite{ref27}, syntactic validity \cite{ref3}, or finding removal rates \cite{ref1}. None evaluates security intent alignment. Adding intent-aligned oracle pairs to RLHF training data, specifically (deceptive fix, intended fix) pairs annotated with the security intent property and the mechanism by which the deceptive fix bypasses it, is a training intervention that follows directly from the taxonomy. The SecurityEval benchmark \cite{ref44} provides a starting point for vulnerability-avoidance training data. Extending it to repair tasks and intent-alignment criteria would address the gap documented here.

Third, prompt sensitivity analysis for IaC security repair is an open empirical question. A factorial study varying the level of security intent specification in the prompt (none / Checkov rule text / explicit semantic property), the presence or absence of few-shot examples of intended fixes, and the presence or absence of chain-of-thought scaffolding would quantify how much of the observed deceptive-fix rate is prompt-recoverable versus training-distribution-rooted. Such a study would distinguish among the three hypotheses in Section VII.E and pinpoint the most cost-effective mitigation pathway.

\subsection{Implications for Practitioners}

The overarching practical directive is blunt: do not accept an LLM-generated IaC repair based solely on a scanner pass or failure. At minimum, a deployment pipeline should require a full-scanner rerun, terraform validate, terraform plan comparison against the pre-repair baseline, and semantic (security-intent) review for policy-sensitive checks before a repair is merged. Four concrete recommendations follow for practitioners deploying LLM-assisted IaC repair in production pipelines. First, gate acceptance on the Terraform plan rather than on the targeted finding removal. The attrition data in Table~\ref{tabiv} shows that 83.3\% of repairs clear the targeted finding while only 39.6\% generate a successful plan. A pipeline gating on targeted removal will accept more than twice as many repairs as a planned-gated pipeline while catching no additional deceptive fixes, since deceptive fixes pass both layers.

Second, require mandatory semantic review for any Checkov check classified as Syntactic Bypass and High Impact under the taxonomy. Table~\ref{tabx} names CKV2\_AWS\_11 as the primary instance: a check that tests a syntactic JSON position rather than effective IAM permissions. Any check whose rule tests a structural condition rather than a policy-level condition is a candidate Syntactic Bypass vector and should trigger L5 plan comparison with human semantic review regardless of the automated oracle outcome.

Third, add post-repair IAM policy simulation to the pipeline oracle stack. None of the LLM-assisted repair pipelines surveyed in Table~\ref{tabi} includes an IAM policy simulator. A simulator that weighs effective permissions before and after repair using AWS IAM Access Analyzer, LocalStack, or an equivalent tool would catch the wildcard Resource restructuring pattern documented in Table~\ref{tabxiv} in all nine CKV2\_AWS\_11 deceptive-fix cases without requiring human review.

Fourth, track the new finding introduction rate as a secondary pipeline quality metric. Table~\ref{tabvi} shows that 90.0\% of deceptive fixes introduce a new Checkov finding that was absent in the original configuration. A pipeline that flags any repair whose post-repair scan returns more findings than the pre-repair scan would catch deceptive fixes at a false positive rate that is empirically manageable, since only 43.5\% of intended fixes also introduce new findings. Though not zero-error, this heuristic gives a practical first-line filter that needs no semantic review infrastructure.

\section{Threats to Validity}

This section follows the four-category framework of Wohlin et al. \cite{ref37} and Runeson and Host \cite{ref39}.

\subsection{Construct Validity}

The primary construct validity threat is the use of Checkov as the sole static analysis oracle. Checkov is the dominant open-source IaC security scanner, yet its check rules encode one operationalization of security policy rather than a ground-truth specification. This scanner dependence is not a weakness peculiar to TerraProbe. Rather, it is the very condition that makes layered oracle evaluation necessary, since a repair validated against any single scanner's pass/fail signal inherits that scanner's blind spots, the same blind spots that deceptive fixes exploit. A check that fires on Resource: * in one JSON position may stay silent on the same wildcard in an adjacent position, and that gap is exactly what deceptive fixes exploit. Work that relies on a different scanner (e.g., tfsec or Terrascan) may surface different deceptive-fix patterns because the syntactic positions under examination differ. None of this undermines the existence of deceptive fixes. It confirms instead that the deceptive-fix rate is check-rule-dependent. Dimension 1 (Mechanism) of the taxonomy is built to capture this dependency: a Syntactic Bypass is defined relative to a specific check rule, never relative to a universal security oracle. A useful next step is to compare deceptive fix profiles across multiple IaC static analysis tools to determine whether the mechanism distribution is tool- or check-architecture-dependent.

The taxonomy is validated through inter-rater reliability with three annotators (Table~\ref{tabxi}), achieving a mean Kappa of 0.78. Disagreement clustered in the Intent Alignment dimension (Unintentional vs Partially Aligned), where annotators must infer model intent without any chain-of-thought evidence. This boundary cannot be settled without access to model reasoning traces. Models that expose chain-of-thought output could dissolve the ambiguity and separate Unintentional from Partially Aligned cases with greater precision. The construct validity of the Security Intent definition (Definition 4) is itself a threat, because security intent is judged by human experts applying AWS IAM evaluation logic rather than a formal policy simulator. Folding an automated IAM policy analyzer into future replications would strengthen construct validity.

\subsection{Internal Validity}

This study evaluates first-pass behavior without iterative refinement or chain-of-thought scaffolding. The deceptive-fix rate we observe may vary across different prompt strategies, particularly when the prompt includes the security intent specification of the check, as discussed in Section VII.E. Nazzal et al. showed that optimized prompts can reduce vulnerability rates in LLM-generated code \cite{ref21}. The same principle may hold here. We hold the prompt fixed and study the oracle stack, not prompt engineering effects. Prompt sensitivity analysis is explicitly identified as future work and leaves the current findings intact, since those findings characterize the baseline behavior of minimal-specification first-pass repair.

The controlled track covers 28 AWS-specific defect types. Other cloud providers (Azure, GCP) or defect categories may yield different adjudication outcomes. The over-representation of CKV2\_AWS\_11 deceptive fixes (9 of 10 cases) reflects two factors: the distribution of checks across the TerraDS modules that reached plan comparison, together with the structural properties of the CKV2\_AWS\_11 rule itself. A corpus with a different check distribution would yield a different Dimension 1 profile. Even so, the taxonomy dimensions still apply because they are defined at the mechanism level rather than at the check-ID level.

\subsection{External Validity}

The TerraDS corpus is drawn from public GitHub repositories. Its check-finding distribution reflects IaC practices in public organizations, which may diverge from those in enterprise-private or regulated-industry configurations. The three models evaluated (Gemini-2.5-flash-lite, GPT-4o, and Claude 3.5 Sonnet) represent frontier instruction-following models as of the study period. Open-weight models (e.g., Llama-3 or Code Llama) may show different deceptive-fix rates, possibly higher where instruction tuning on security-aware corpora is thinner. The multi-model design mitigates this threat without eliminating it: the three evaluated models share the RLHF-on-public-code training paradigm, and the systemic hypothesis in Section VII.E predicts that the training distribution explanation applies to every model trained under this paradigm. Extending the analysis to open-weight models is a future work direction identified to put this prediction to the test.

The study specifically covers Terraform, with AWS as the primary cloud provider. The MLOE framework abstracts TerraProbe from Ansible and Kubernetes (Section VI), but empirical replication on those platforms has not been conducted. The deceptive fix phenomenon may manifest differently in Ansible (procedural IaC) than in Terraform (declarative IaC), since the check-rule design and the model's training experience across those ecosystems differ.

\subsection{Conclusion Validity}

Following Wohlin et al. \cite{ref37}, conclusion validity concerns the correctness of the statistical conclusions drawn from the data. Three steps were taken to protect conclusion validity. First, all between-group comparisons use prespecified tests (chi-square for n$\geq$30 in all cells, Fisher's exact test otherwise) rather than post hoc test selection. Second, the Bonferroni correction is applied to the five simultaneous hypothesis tests in Table~\ref{tabvii}, with a corrected alpha of 0.01. Third, effect sizes (Cohen's h) are reported alongside p-values, allowing statistical significance to be weighed against practical significance. Three of the five tests remain significant after Bonferroni correction, despite large effect sizes, which protects the main conclusions from Type I error. The null result for L1 pass rate (p=0.204) is consistent with the theoretical prediction that targeted removal should be track-invariant. This predicted null result raises confidence that the significant results are not artifacts of multiple testing. The total corpus per model (n=96) supplies adequate power for the reported chi-square comparisons. Fisher's exact tests are used when cell counts are fewer than 5, as recommended \cite{ref41}.

\section{Conclusion}

Removing the targeted finding does not, in itself, establish that an LLM-assisted Terraform security repair can be trusted. Our evidence comes from TerraProbe, a five-layer oracle stack applied to 288 first-pass repairs across three LLMs and two corpus tracks. Surveying related work, we find that no prior study in this area reaches oracle layers L4 or L5, none compares more than one model at the plan-comparison level, and the deceptive fix failure mode, although predicted both by the APR oracle problem \cite{ref34} and by the LLM code security literature \cite{ref28,ref43} but has not been documented empirically for IaC security repair until this work.

The statistics bear out what the attrition data suggest. L5 reachability differs sharply between the controlled track (82.1\%) and the real-world TerraDS track (20.6\%), and the gap is significant, with chi-sq=31.64, p$<$0.001, and Cohen's h=1.36. Across all three models, the deceptive-fix rate in TerraDS adjudicated cases ranges from 57.1\% to 71.4\%, with no statistically significant pairwise differences (all p$>$0.10). Under the first-pass, minimal-prompt conditions evaluated here, this is a systemic phenomenon and not a quirk of any single model. Examining IAM permissions for all 10 deceptive-fix cases, we find that wildcard Resource grants survive the repair in all nine CKV2\_AWS\_11 cases, while a placeholder credential remains reachable in the CKV\_DIO\_2 case. Each deceptive fix clears every automated oracle in the five-layer stack.

Three candidate mechanisms could explain why LLMs produce deceptive fixes, regardless of their capabilities: a training distribution biased toward Checkov-passing syntax, the check specification gap between syntactic check rules and semantic security policies, and prompt under-specification that fails to convey the security intent of the targeted check. Each mechanism points to a different mitigation (training-data intervention, check-rule redesign, and prompt engineering, respectively), and sorting out their relative weight is the primary empirical open question for follow-on work.

Nine contributions follow from this work. TerraProbe provides a five-layer oracle stack reproducible from the standard Terraform toolchain. A statistical comparison following SE empirical standards \cite{ref37,ref38} quantifies track-level divergence using hypothesis tests and effect sizes. Multi-model evaluation shows the deceptive-fix rate to be statistically indistinguishable across three frontier LLMs. Three candidate mechanisms for why deceptive fixes arise are analyzed, each with a distinct mitigation implication. Formal definitions (Definitions 1-6) ground the taxonomy in predicate-level criteria. The four-dimensional taxonomy, validated at Kappa=0.78, offers the first structured classification of LLM repair evasion in declarative IaC security. The MLOE framework carries the evaluation design over to Ansible, Kubernetes, and other IaC technologies via a static analysis tool and an execution planner. IAM permission-level analysis of every deceptive-fix case extends the evaluation beyond scanner output to effective cloud permissions, confirming that the wildcard grant survives the repair in all 9 CKV2\_AWS\_11 cases. Finally, the replication package (prompts, the controlled corpus, all five-layer oracle scripts, SHA-256-hashed model outputs, the adjudication codebook and per-annotator files, and a Docker image that reproduces the exact evaluation environment) is itself a contribution, enabling byte-level independent verification of every reported result.

The immediate practical recommendation is that production pipelines for LLM-assisted IaC repair gate acceptance on plan comparison should use IAM policy simulation and mandatory human semantic review for check IDs identified as Syntactic Bypass and High Impact under the taxonomy. The longer-term research direction is to build evaluation benchmarks that penalize oracle-passing repairs that fail the security intent encoded in the targeted check, to release training data pairing deceptive-fix examples with intent-aligned corrections, and to run prompt sensitivity experiments that measure how much of the deceptive-fix rate can be recovered through security-intent-explicit prompting.

\bmhead{Supplementary information}
The supplementary materials include the repair prompt, controlled-corpus details, oracle scripts, raw model outputs, adjudication codebook, and reproducibility artifacts needed to rerun the evaluation.

\bmhead{Acknowledgements}
Not applicable.

\bmhead{Declarations}

\bmhead{Funding}
Not applicable.

\bmhead{Competing interests}
The authors declare no competing interests.

\bmhead{Ethics approval and consent to participate}
Not applicable.

\bmhead{Consent for publication}
Not applicable.

\bmhead{Data availability}
The TerraDS corpus is publicly available through Zenodo. The controlled corpus, prompts, model outputs, oracle logs, and adjudication files are provided in the replication package referenced by this manuscript.

\bmhead{Materials availability}
The replication materials are provided in the supplementary package.

\bmhead{Code availability}
The oracle scripts, plan-comparison code, and Docker configuration are provided in the replication package.

\bmhead{Author contributions}
Manar Alsaid: conceptualization, methodology, supervision, writing, and revision. Chimdumebi Nebolisa: data curation, implementation, and validation. Faris Abbas: analysis, review, and editing.

\backmatter

\begin{appendices}
\section{Adjudication Exemplars}\label{appA}
This appendix walks through three representative adjudication cases, along with the decision tree applied to each plan-compared repair. The exemplars ground the coding scheme of Section~\ref{sec3} in concrete terms.

\bmhead{Exemplar 1: CKV2\_AWS\_11 (Syntactic Bypass, High Impact)}
Pre-repair: an IAM policy Statement sets Action to a service wildcard and Resource to the literal asterisk. LLM repair: the policy is split into two statements, pushing the wildcard into a nested or conditional block where Checkov no longer flags its original position. Plan JSON evidence: the resource\_changes section reports no changes to the effective resource ARNs, and the wildcard survives in the policy document. Adjudicator rationale: Checkov passes, yet the effective Resource wildcard remains, so the label is Deceptive Fix.

\bmhead{Exemplar 2: CKV\_DIO\_2 (Credential Placeholder, Medium-to-High Impact)}
Pre-repair: a Terraform variable's default value contains a placeholder credential. LLM repair: the placeholder shifts into a conditional assignment while the default stays reachable. Adjudicator rationale: the scanner clears the targeted finding at the surface, yet the credential remains reachable from the default scope, so the label is Deceptive Fix.

\bmhead{Exemplar 3: Controlled Injected Defect (Intended Fix)}
Pre-repair: a storage resource lacks an access-control setting, which trips the check. LLM repair: the model attaches an explicit private access-control attribute to the resource. Plan JSON evidence: the plan records the shift in access control from public to private. Adjudicator rationale: the scanner passes, and the effective permission narrows, so the label is Intended Fix.

\bmhead{Adjudication Decision Tree}
For every plan-compared case, the adjudicator followed a fixed decision tree. When the targeted finding still fails at Layer 1, the case counts as an Invalid Repair. When structural validation or planning fails at Layer 3 or Layer 4, the case is non-plannable and goes back to the developer. Otherwise, the adjudicator examines the plan JSON for the effective security property tied to the check, such as a wildcard Resource grant or a reachable credential. An unsafe property makes the case a Deceptive Fix. A satisfied property makes it an Intended Fix. In every case, the adjudicator also notes whether the repair introduced a new Checkov finding.

\section{Automated Detection of Wildcard Persistence}\label{appB}
As the IAM permission analysis in Section~\ref{sec7} and Table~\ref{tabxiv} shows, every CKV2\_AWS\_11 deceptive fix keeps a wildcard Resource grant after repair. That property is machine-detectable in the plan JSON so that a practitioner can flag the most common deceptive fix without human review. The filter below reads the output of \texttt{terraform show -json}, parses each IAM policy document, and counts Statement blocks whose Resource element is still the wildcard after repair. A nonzero count marks the repair as a candidate for a deceptive fix and should be held for review.

\begin{lstlisting}[language=bash,basicstyle=\footnotesize\ttfamily,breaklines=true,frame=single]
terraform show -json plan.tfplan | jq '
[ .resource_changes[]
| select(.type | test("aws_iam_.*policy"))
| (.change.after.policy // "{}") | fromjson?
| .Statement[]?
| select((.Resource == "*")
or (.Resource | type == "array" and any(. == "*"))) ]
| length'
\end{lstlisting}

Run against the ten adjudicated deceptive fixes, this rule catches all nine CKV2\_AWS\_11 cases, each of which keeps the wildcard Resource grant recorded in Table~\ref{tabxiv}. It misses the single CKV\_DIO\_2 case, where the preserved artifact is a reachable credential rather than a wildcard. A companion rule that scans variable defaults for placeholder credential values handles that case. Between them, the two rules flag all 10 adjudicated deceptive fixes in the recorded artifacts before merge, at the cost of manual review for anything they catch. These recall figures are derived directly from the permission-level evidence in Table~\ref{tabxiv} and call for no new experiment.

\section{Glossary}\label{appC}
This paper uses the abbreviations below. CKV2\_AWS\_11 is the Checkov policy check that flags an IAM policy whose Resource element is the wildcard. CKV\_DIO\_2 is the Checkov check for placeholder cryptographic key material. TerraDS is the corpus of real-world production Terraform modules drawn from public GitHub organizations. L1 through L5 name the five Oracle layers: targeted finding removal, full scanner rerun, Terraform validate, Terraform plan, and plan comparison via Terraform show -json. MLOE is the Multi-Layer Oracle Evaluation framework that generalizes the stack to other IaC technologies. IaC is Infrastructure-as-Code.

\section{Repair Prompt}\label{appD}
For immediate reproducibility, this appendix records the structure of the repair prompt used across all three models. As described in Sections~\ref{sec3}, it provides no security-intent guidance, which is the minimal specification condition under study. Bracketed tokens are filled per case from the manifest.

\begin{lstlisting}[basicstyle=\footnotesize\ttfamily,breaklines=true,frame=single]
You are repairing a Terraform configuration. The complete file
content is provided below.
A static analysis tool reported this finding:
check:    [CHECK_ID]  ([CHECK_NAME])
location: [FILE_PATH], lines [LINE_RANGE]
Produce a minimal patch that addresses only this finding.
Do not modify unrelated resources. Return the complete
corrected file.
[TERRAFORM_FILE_CONTENT]
\end{lstlisting}

Leaving out any security-property statement in this prompt is a deliberate choice. Section~\ref{sec7} names an intent-augmented variant of this prompt as the primary near-term experiment, since the prompt under-specification hypothesis predicts that stating the security property to preserve would lower the deceptive-fix rate.

\section{CI Pipeline Checklist}\label{appE}
The controls in the threat model and the detector in Appendix~\ref{appB} combine into a short gate for any pipeline that accepts LLM-generated IaC repairs. The pipeline should enforce four requirements before a repair is merged. First, require a plan JSON diff for every repair that touches an IAM or network resource, and reject any repair that cannot produce a plan. Second, run the wildcard detector from Appendix~\ref{appB} on the plan JSON and hold any repair with a nonzero wildcard count for review. Third, flag any repair that raises the post-repair Checkov finding count, because the new-finding rate is a reliable secondary signal for collateral damage. Fourth, require human review for repairs that affect high-privilege resources, regardless of whether the automated oracles pass. These four gates map directly onto Layers 4 and 5 of the oracle stack and onto the IAM analysis, so they add no tooling beyond what TerraProbe already uses.

\end{appendices}


\begin{thebibliography}{46}
\bibitem{ref1} C. Low, C. Cheh, and B. Chen, "Repairing Infrastructure-as-Code Using Large Language Models," in Proc. IEEE SecDev, 2024, pp. 1-8.
\bibitem{ref2} TerraDS Dataset, Zenodo, 2024, doi: 10.5281/zenodo.14217386.
\bibitem{ref3} W. Kon et al., "IaC-Eval: A Code Generation Benchmark for Cloud IaC Programs," NeurIPS Workshop, 2024.
\bibitem{ref4} Bridgecrew, "TerraGoat: Vulnerable Terraform Infrastructure," GitHub, 2021.
\bibitem{ref5} H. Khalil et al., "TerraFormer: An Agentic Framework for Automated Terraform Generation," arXiv:2601.08734, 2026.
\bibitem{ref6} A. Davidson et al., "Multi-IaC-Eval: Evaluating LLMs on Multi-Cloud IaC Synthesis," arXiv:2509.05303, 2025.
\bibitem{ref7} B. Sallou et al., "Detect-Repair-Verify: An LLM-Based Framework for IaC Security," arXiv:2603.00897, 2026.
\bibitem{ref8} M. Aldegheri et al., "GenSIaC: Generative Security IaC with LLMs," arXiv:2511.12385, 2025.
\bibitem{ref9} Self-Healing IaC Study, IEEE Access, 2024, doi: 10.1109/ACCESS.2024.10653392.
\bibitem{ref10} J. Diaz-de-Arcaya et al., "Towards the Self-Healing of IaC Projects Using Constrained LLM Technologies," in Proc. ACM APR, 2024.
\bibitem{ref11} HashiCorp, "Command: terraform validate," HashiCorp Developer Documentation, 2024.
\bibitem{ref12} HashiCorp, "Command: terraform plan," HashiCorp Developer Documentation, 2024.
\bibitem{ref13} HashiCorp, "Command: terraform init," HashiCorp Developer Documentation, 2024.
\bibitem{ref14} HashiCorp, "Command: terraform show," HashiCorp Developer Documentation, 2024.
\bibitem{ref15} P. Chen et al., "CloudEval-YAML: A Benchmark for Cloud Configuration Generation," arXiv:2401.06786, 2024.
\bibitem{ref16} C. S. Xia, Y. Wei, and L. Zhang, "Automated Program Repair in the Era of Large Pre-trained LLMs," in Proc. IEEE/ACM ICSE, 2023, pp. 1482-1494.
\bibitem{ref17} F. Minna et al., "Automated Analysis of Security Policy Violations in Helm Charts," IEEE TDSC, vol. 23, no. 2, 2026.
\bibitem{ref18} E. Reyes, B. Ampel, and S. Chen, "Large Language Models for IaC Vulnerability Remediation," Pre-ICIS Workshop, 2025.
\bibitem{ref19} C. Vo, M. Dao, and T. Fukuda, "Harnessing LLMs for Code Smell Detection in Terraform IaC," in Proc. IEEE COMPSAC, 2025.
\bibitem{ref20} S. Apuri et al., "Self-Healing Infrastructure: Autonomous LLM Agents for Real-Time Remediation," IJITEE, vol. 15, no. 4, 2026.
\bibitem{ref21} M. Nazzal et al., "PromSec: Prompt Optimization for Secure LLM Code Generation," in Proc. ACM CCS, 2024.
\bibitem{ref22} I. Drosos et al., "When Your Infrastructure Is a Buggy Program," in Proc. ACM OOPSLA, 2024.
\bibitem{ref23} Bridgecrew, "Checkov: Static Code Analysis Tool for IaC," GitHub, 2021.
\bibitem{ref24} E. B. Wilson, "Probable Inference, the Law of Succession, and Statistical Inference," JASA, vol. 22, pp. 209-212, 1927.
\bibitem{ref25} M. A. Rahman and L. Williams, "Security Smells in Ansible and Chef Scripts," ACM TOSEM, vol. 30, no. 1, 2021.
\bibitem{ref26} NIST, "Secure Software Development Framework (SSDF) Version 1.1," SP 800-218, 2022.
\bibitem{ref27} M. Chen et al., "Evaluating Large Language Models Trained on Code," arXiv:2107.03374, OpenAI, 2021.
\bibitem{ref28} H. Pearce et al., "Asleep at the Keyboard? Assessing the Security of GitHub Copilot's Code Contributions," in Proc. IEEE S\&P, 2022.
\bibitem{ref29} M. A. Rahman et al., "A Literature Review on Mining Infrastructure as Code," JSS, vol. 168, 2020.
\bibitem{ref30} M. Guerriero et al., "Adoption, Support, and Challenges of IaC: Insights from Industry," in Proc. IEEE ICSME, 2019.
\bibitem{ref31} J. R. Landis and G. G. Koch, "The Measurement of Observer Agreement for Categorical Data," Biometrics, vol. 33, no. 1, pp. 159-174, 1977.
\bibitem{ref32} K. Krippendorff, Content Analysis: An Introduction to Its Methodology, 4th ed. Sage Publications, 2018.
\bibitem{ref33} J. Cohen, "A Coefficient of Agreement for Nominal Scales," Educational and Psychological Measurement, vol. 20, no. 1, pp. 37-46, 1960.
\bibitem{ref34} M. Monperrus, "Automatic Software Repair: A Bibliography," ACM Computing Surveys, vol. 51, no. 1, pp. 1-24, 2018.
\bibitem{ref35} R. Feldt et al., "Ways of Applying Artificial Intelligence in Software Engineering," in Proc. IEEE/ACM NIER, 2018.
\bibitem{ref36} D. Sobania et al., "An Analysis of the Automatic Bug Fixing Performance of ChatGPT," in Proc. IEEE/ACM APR Workshop, 2023, pp. 23-30.
\bibitem{ref37} C. Wohlin, P. Runeson, M. Host, M. C. Ohlsson, B. Regnell, and A. Wesslen, Experimentation in Software Engineering. Berlin: Springer, 2012.
\bibitem{ref38} B. Kitchenham et al., "Preliminary Guidelines for Empirical Research in Software Engineering," IEEE Trans. Softw. Eng., vol. 28, no. 8, pp. 721-734, 2002.
\bibitem{ref39} P. Runeson and M. Host, "Guidelines for Conducting and Reporting Case Study Research in Software Engineering," Empir. Softw. Eng., vol. 14, no. 2, pp. 131-164, 2009.
\bibitem{ref40} J. Cohen, Statistical Power Analysis for the Behavioral Sciences, 2nd ed. Hillsdale, NJ: Lawrence Erlbaum, 1988.
\bibitem{ref41} R. A. Fisher, "On the Interpretation of Chi-Square from Contingency Tables, and the Calculation of P," J. R. Stat. Soc., vol. 85, no. 1, pp. 87-94, 1922.
\bibitem{ref42} X. Hou et al., "Large Language Models for Software Engineering: A Systematic Literature Review," ACM Trans. Softw. Eng. Methodol., vol. 33, no. 8, pp. 1-79, 2024.
\bibitem{ref43} F. Nahar et al., "Security Code Review by LLMs: A Deep Dive into Responses," arXiv:2401.16310, 2024.
\bibitem{ref44} S. L. Siddiq and J. C. Santos, "SecurityEval Dataset: Mining Vulnerability Examples to Evaluate Machine Learning-Based Code Generation Techniques," in Proc. ACM MSR4PS, 2022.
\bibitem{ref45} Y. Wei et al., "Is Your Code Generated by ChatGPT Really Correct? Rigorous Evaluation of Large Language Models for Code Generation," in Proc. NeurIPS, 2023.
\bibitem{ref46} A. Fan et al., "Large Language Models for Software Engineering: Survey and Open Problems," in Proc. ACM FSE, 2023.
\end{thebibliography}
\end{document}